\documentclass[lettersize,journal]{IEEEtran}
\usepackage{amsmath,amsfonts}
\usepackage{algorithm}
\usepackage{array}
\usepackage[caption=false,font=normalsize,labelfont=sf,textfont=sf]{subfig}
\usepackage{textcomp}
\usepackage{stfloats}
\usepackage{url}
\usepackage{verbatim}
\usepackage{graphicx}
\usepackage{arydshln} 
\usepackage{cite}
\usepackage{amssymb}
\usepackage{colortbl}
\usepackage{booktabs}
\usepackage{pifont}
\usepackage{multirow}
\usepackage{multicol}
\usepackage{subcaption}
\usepackage{algpseudocode}
\begin{document}

\title{Rethinking Knowledge Retrieval for Generation: A Survey on RAG Architectures and Applications}

\author{
Meghana Sunil$^{*}$, 
Shravya V$^{*}$, 
Shravan Venkatraman, 
Joe Dhanith P R
\\
School of Computer Science and Engineering, Vellore Institute of Technology, Chennai, India
\thanks{$^{*}$ Equal contribution.}
}

\markboth{Journal of \LaTeX\ Class Files,~Vol.~14, No.~8, August~2021}%
{Shell \MakeLowercase{{et al.}}: A Sample Article Using IEEEtran.cls for IEEE Journals}


\maketitle

\begin{abstract}
Large Language Models (LLMs) have demonstrated remarkable fluency and versatility across natural language tasks but remain fundamentally limited by their static knowledge and susceptibility to hallucinations, especially in domains requiring up-to-date or attribute-grounded information. Retrieval-Augmented Generation (RAG) addresses these challenges by integrating external retrieval mechanisms with generative models, enabling dynamic, context-aware generation grounded in verifiable data sources. This survey presents a comprehensive examination of RAG as a modular and evolving paradigm that enhances factual reliability, adaptability, and task alignment in LLM-based systems. We formalize the RAG framework through its three foundational components - retrieval, generation, and augmentation - and survey state-of-the-art methods spanning dense and sparse retrievers, fusion strategies, embedding optimizations, and reinforcement learning-based retrieval policies. Anchored around four emerging axes - efficiency, security, user-centric interactivity, and complex reasoning - we categorize recent innovations and highlight their implications for scalability, robustness, and personalization. The paper also reviews advances in evaluation protocols, domain-specific applications, and architectural variants such as Naïve RAG, Advanced RAG, and Modular RAG. Finally, we identify persistent challenges and outline future directions aimed at advancing the integration of retrieval with LLMs for more grounded, interpretable, and controllable generation.
\end{abstract}


\section{Introduction}
Large Language Models (LLMs) have achieved remarkable progress in natural language generation, powering applications ranging from conversational agents to complex summarization and question answering systems. However, a persistent limitation of these models lies in their reliance on static, pre-trained parameters. As a result, LLMs often struggle to incorporate new or domain-specific information and are prone to generating inaccurate or hallucinated content - particularly in scenarios with sparse or attribute-based inputs~\cite{li2025enhancingretrievalaugmentedgenerationstudy}.

These challenges become critical in knowledge-intensive tasks such as review generation, dialogue systems, and data-to-text (D2T) generation, where factual accuracy and contextual relevance are paramount~\cite{krishna2025factfetchreasonunified}. Retrieval Augmented Generation (RAG) offers a compelling solution to these limitations by coupling LLMs with external knowledge retrieval systems. Instead of generating responses solely based on internalized knowledge, RAG models dynamically fetch semantically relevant information from structured or unstructured sources - such as relational databases, RDF graphs, and textual corpora - at inference time~\cite{singh2025agenticretrievalaugmentedgenerationsurvey, han2025retrievalaugmentedgenerationgraphsgraphrag}.

This retrieval mechanism allows the generation process to be grounded in verifiable external data, thereby enhancing factual consistency and enabling real-time adaptation to evolving information~\cite{wang2025chainofretrievalaugmentedgeneration}. In personalized applications like product reviews or recommendation systems, RAG architectures can retrieve user- or item-specific information to tailor responses more effectively. Moreover, the integration of reinforcement learning has enabled dynamic optimization of retrieval quality, further improving the relevance and efficiency of augmented content~\cite{zhou2025trustragenhancingrobustnesstrustworthiness}.

Formally, RAG is defined as a hybrid framework that enhances the response generation capabilities of LLMs by incorporating retrieval-based augmentation. It consists of three core components: retrieval, where the input query is used to fetch relevant document chunks based on learned embeddings; generation, where the retrieved context is fused with the query to produce a coherent response; and augmentation, where enhancements such as better indexing strategies, fine-tuning techniques, and embedding refinements are employed to improve end-to-end performance. This modularity allows RAG to be adapted across a wide range of tasks and domains, including medical diagnosis, educational tools, and customer support systems - areas where factual grounding is non-negotiable~\cite{jeong2025videoragretrievalaugmentedgenerationvideo}.

Recent developments have expanded the design space of RAG, giving rise to variants such as Naïve RAG, Advanced RAG, and Modular RAG, each introducing targeted improvements in retrieval precision, integration depth, and generation quality~\cite{fan2025miniragextremelysimpleretrievalaugmented}. Evaluation methodologies have also evolved, with task-specific benchmarks and metrics now assessing both the informativeness of retrieved passages and the quality of generated outputs~\cite{retrieval-strategies2024}.

As research continues to refine both the theoretical and practical underpinnings of RAG, its role has expanded beyond inference-time augmentation to deeper integration with LLM training pipelines, showcasing its versatility, scalability, and potential to mitigate some of the most pressing limitations of modern language models~\cite{liang2025saferagbenchmarkingsecurityretrievalaugmented}.

This survey investigates the evolving landscape of RAG by organizing recent advancements into four major themes: efficiency, security, interactivity, and complex reasoning. Each theme is defined by a set of central research questions and representative contributions that address practical and theoretical challenges in the design and deployment of RAG-based systems.

\begin{enumerate}
    \item \textbf{Compression and Efficiency-Driven RAG.} As LLMs interface with increasingly large external knowledge sources, optimizing retrieval efficiency while preserving output quality has become a central challenge~\cite{fan2025miniragextremelysimpleretrievalaugmented}. Key research questions include: How can retrieval be compressed without degrading accuracy? How can training data be selected more effectively? What trade-offs exist between minimal-context and full-dataset retrieval? Contributions in this area include large-scale retrieval techniques that scale to billions of documents, few-shot retrieval strategies to reduce inference overhead, and paradigms that rely on curated training data in place of external corpora.

    \item \textbf{Defensive RAG: Addressing Bias, Vulnerabilities, and Privacy.} Given the growing use of RAG in high-stakes applications, concerns around adversarial inputs, contextual bias, and privacy leakage have gained prominence~\cite{liang2025saferagbenchmarkingsecurityretrievalaugmented, zhou2025trustragenhancingrobustnesstrustworthiness}. Research questions focus on how to defend against misleading context, mitigate embedded biases, and design privacy-preserving retrieval mechanisms. Notable contributions include reinforcement learning approaches for ethical alignment, evaluations that expose retrieval-induced bias, and architectures that obfuscate sensitive information during retrieval and generation.

    \item \textbf{Interactive, User-Centric RAG Systems.} Personalization and adaptability are critical for user-facing systems. This category addresses research questions around dynamic retrieval strategies, the role of structured knowledge (e.g., knowledge graphs) in interactive pipelines, and how RAG can power adaptive recommendation systems~\cite{han2025retrievalaugmentedgenerationgraphsgraphrag, singh2025agenticretrievalaugmentedgenerationsurvey}. Key contributions include personalized retrieval mechanisms that evolve with user input, graph-based query refinement techniques, and open-domain recommendation frameworks enhanced by retrieval.

    \item \textbf{RAG for Complex Reasoning and Multi-Step Tasks.} Many real-world scenarios require long-context reasoning and multi-hop inference that exceed the capabilities of standard LLMs. Research in this category explores how retrieval can support iterative reasoning, facilitate multi-step generation, and improve long-context retention~\cite{wang2025chainofretrievalaugmentedgeneration, krishna2025factfetchreasonunified, retrieval-strategies2024}. Contributions span retrieval-augmented knowledge-intensive NLP tasks, multi-turn conversational agents, and memory-efficient designs for handling extended input contexts.
\end{enumerate}

This paper is structured to offer a comprehensive and systematic review of RAG in the context of LLMs. It begins with the present introduction, which outlines the motivation, scope, key questions, and structure of the survey. Section~\ref{sec:rag_foundations} introduces the foundational principles of RAG, covering its conceptual formulation, theoretical underpinnings, and historical evolution from early memory-based models to modern retrieval-integrated architectures.

Section~\ref{sec:taxonomy} presents a detailed taxonomy of RAG-LLM systems, categorizing them based on retrieval methods (dense, sparse, hybrid), generation models (decoder-only, encoder-decoder), integration strategies (early fusion, late fusion, mixed pipelines), and extensions to multi-modal settings. This taxonomy highlights the architectural diversity and design trade-offs across existing approaches.

Section~\ref{sec:applications} surveys real-world applications of RAG-LLMs, including use cases in scientific literature processing, conversational agents, code generation, enterprise knowledge systems, and open-domain question answering. Section~\ref{sec:limitations} discusses major limitations and challenges faced by current RAG systems, such as retrieval bottlenecks, hallucination propagation, domain mismatch, scalability issues, and lack of interpretability.

Section~\ref{sec:future_work} outlines emerging trends and promising directions for future work, including reinforcement learning-based retrieval policies, neuro-symbolic integration, and multi-modal retrieval pipelines. Finally, Section~\ref{sec:conclusion} summarizes the key insights and reflects on the broader implications of RAG in advancing the factual reliability and adaptability of LLMs.

\begin{figure*}[t] 
    \centering
    \includegraphics[width=0.6\textwidth]{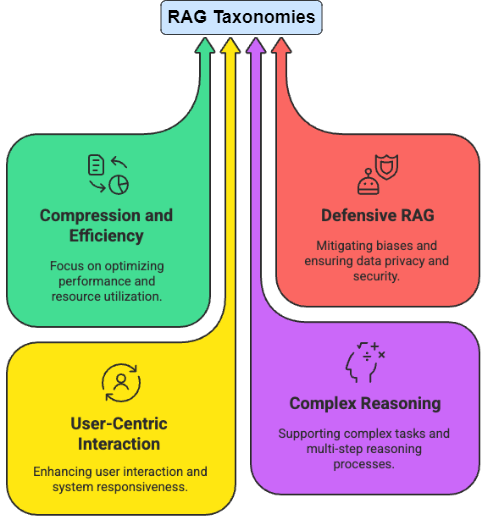} 
    \caption{Overview of the four core taxonomic dimensions of Retrieval-Augmented Generation (RAG) systems explored in this review: {Compression and Efficiency}, {Defensive RAG}, {User-Centric Interaction}, and {Complex Reasoning}. These axes reflect emerging priorities in optimizing performance, enhancing security and fairness, improving user alignment, and enabling multi-step, structured reasoning.}
    \label{fig:classifications} 
\end{figure*}

\section{Foundations of RAG}
\label{sec:rag_foundations}

\subsection{Conceptual Overview}

RAG is a hybrid framework that enhances the capabilities ofLarge Language Models (LLMs) by integrating real-time information retrieval into the text generation process. Unlike conventional LLMs that generate responses solely based on knowledge encoded in their model parameters during pre-training, RAG introduces an external retrieval step to access up-to-date and contextually relevant information from knowledge bases, databases, or document repositories. This mechanism significantly improves factual consistency, reduces hallucinations, and enables the model to remain responsive to dynamic information needs.

The RAG framework operates in two principal stages. In the retrieval stage, the system identifies relevant documents or passages corresponding to a given input query. Retrieval methods include dense approaches - such as Dense Passage Retrieval (DPR), which employs learned embeddings to capture semantic similarity - and sparse techniques like BM25, which rely on term-frequency and inverse-document-frequency scoring. Following retrieval, the generation stage involves conditioning a pre-trained language model (e.g., GPT or T5) on the query along with the retrieved content to synthesize a coherent and factually grounded response.

One of the central advantages of RAG lies in its adaptability. While traditional LLMs require computationally intensive retraining to incorporate new knowledge, RAG systems dynamically fetch relevant information at inference time, allowing for rapid integration of emerging facts. This retrieval-based approach also enhances interpretability by enabling users to trace generated content back to specific source documents, thus improving transparency and trustworthiness. By bridging static parametric knowledge with dynamic external retrieval, RAG offers a scalable and explainable pathway for advancing real-world language understanding and generation systems.

\subsection{How RAG Differs from Standard LLMs}

Traditional LLMs operate under a fixed knowledge paradigm, where the entirety of their world understanding is encoded during pre-training. As a result, these models are constrained by the static nature of their training data and cannot adapt to new information without extensive retraining. This limitation is particularly problematic in domains such as medicine, finance, and current affairs, where factual updates occur frequently. Moreover, standard LLMs do not consult any external reference during inference, increasing the likelihood of hallucinations - plausible-sounding but factually incorrect outputs - and reducing transparency in knowledge provenance.

RAG departs from this architecture by incorporating a retrieval module that fetches relevant external content at inference time. This retrieved context is then used to guide the generation process, ensuring that responses are informed by verifiable sources rather than solely by pre-trained weights. By grounding responses in retrieved documents, RAG mitigates hallucinations and enhances factual accuracy. Additionally, this structure allows users to inspect the origin of each generated response, fostering interpretability and user confidence.

Another core distinction is that RAG adopts a hybrid approach to knowledge representation. It balances parametric knowledge stored within the LLM and non-parametric knowledge obtained via retrieval. This enables the system to retain broad generalization capabilities while remaining adaptable to task-specific or time-sensitive contexts. As a result, RAG is particularly well-suited to applications such as open-domain question answering, task-oriented dialogue systems, and knowledge-intensive NLP pipelines, where response accuracy and source transparency are essential.

\subsection{Theoretical Background}

The shift from purely parametric language models to retrieval-augmented architectures represents a foundational advancement in natural language processing. While conventional LLMs store all their knowledge within learned parameters, RAG systems supplement this internal representation with external information sources, accessed dynamically at inference time. This architectural divergence enables RAG to remain responsive to evolving domains such as healthcare, legal reasoning, and technical support, where factual accuracy and domain specificity are critical.

The retrieval component in RAG plays a central role in this shift. Rather than generating responses solely from internalized representations, RAG actively searches for relevant documents from large-scale knowledge bases, web corpora, or structured databases. Retrieval methods typically fall into two categories: sparse retrieval techniques such as BM25, which rely on term-frequency-based keyword matching, and dense retrieval techniques like Dense Passage Retrieval (DPR), which use neural embeddings to identify semantically similar passages. The retrieved context is then fused with the original input during generation, resulting in responses that are both grounded and explainable.

This hybridization of parametric and non-parametric knowledge addresses a key limitation of standard LLMs - their inability to access or incorporate new information post-training. While parametric models like GPT offer fluent and general-purpose generation, they are inherently static and lack mechanisms for real-time adaptation. In contrast, RAG systems enhance adaptability by retrieving new information as needed, reducing the necessity for retraining and improving factual consistency. Emerging hybrid approaches further refine this balance, combining the fluency and generalization capabilities of parametric models with the contextual precision of retrieval-based augmentation.

\begin{figure*}[htbp]
    \centering
    \includegraphics[width=0.85\linewidth]{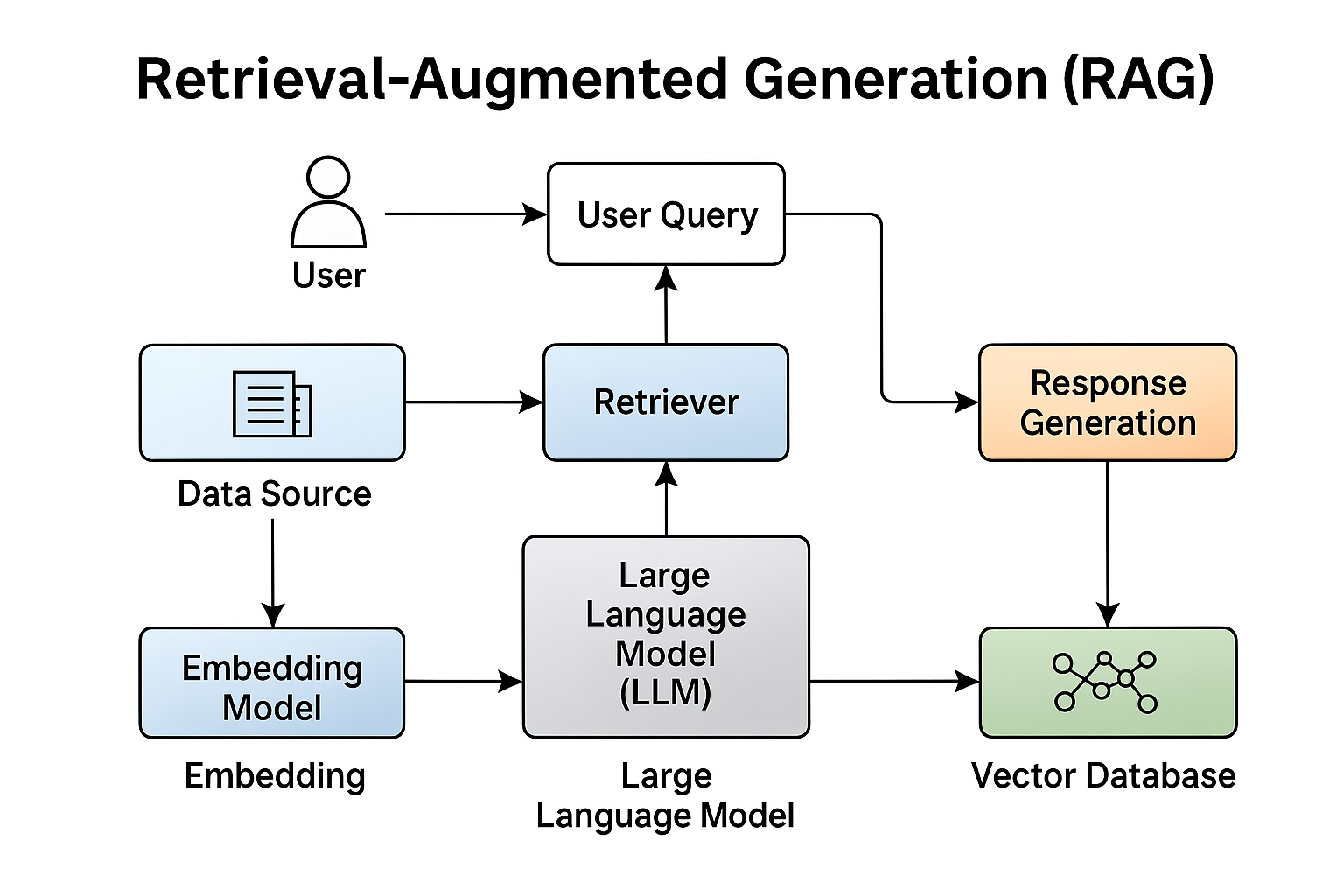}
    \caption{System architecture of a standard Retrieval-Augmented Generation (RAG) pipeline. The user query is encoded and used to retrieve semantically relevant documents from a vector database via a retriever module. These documents are then passed into a large language model alongside the query to generate a grounded response. An embedding model encodes the data corpus offline, and document vectors are stored in a searchable index.}
    \label{fig:rag-architecture}
\end{figure*}

In traditional parametric models, a language model $P_\theta(y \mid x)$ is learned purely from training data, where:
\begin{equation}
P_\theta(y \mid x) = \text{Decoder}(x; \theta)
\end{equation}
Here, $x$ is the input, $y$ is the output, and $\theta$ represents the fixed parameters learned during training.

In Retrieval-Augmented Generation, this paradigm is extended by incorporating an external retrieval mechanism. Given a query $x$, a set of relevant documents $D = \{d_1, d_2, \dots, d_k\}$ is retrieved from an external corpus $\mathcal{C}$ using a retriever function $R$:
\begin{equation}
D = R(x, \mathcal{C})
\end{equation}

The generation probability is then reformulated as:
\begin{equation}
P(y \mid x) = \sum_{d \in D} P(y \mid x, d) \cdot P(d \mid x)
\end{equation}
where:
\begin{itemize}
    \item $P(d \mid x)$ is the probability of retrieving document $d$ given query $x$.
    \item $P(y \mid x, d)$ is the generation probability conditioned on both the query and the retrieved document.
\end{itemize}

\paragraph{Retrieval Models.}
\begin{itemize}
    \item \textbf{Sparse Retrieval (e.g., BM25)} employs a TF-IDF-based scoring mechanism:
    \begin{multline}
    \text{Score}_{\text{BM25}}(x, d) = \sum_{t \in x \cap d} \text{IDF}(t) \cdot \\
    \frac{f(t, d) \cdot (k+1)}{f(t, d) + k \cdot \left(1 - b + b \cdot \frac{|d|}{\text{avgdl}}\right)}
    \end{multline}
    where $f(t, d)$ is the frequency of term $t$ in document $d$, and $k$, $b$ are hyperparameters controlling term frequency scaling and document length normalization.
    
    \item \textbf{Dense Retrieval (e.g., DPR)} relies on neural embeddings and computes a similarity score via the dot product:
    \begin{equation}
    \text{Score}_{\text{DPR}}(x, d) = \phi_q(x)^\top \phi_d(d)
    \end{equation}
    where $\phi_q$ and $\phi_d$ are learned neural encoders for queries and documents, respectively.
\end{itemize}

The retrieval distribution is typically normalized using a softmax:
\begin{equation}
P(d \mid x) = \frac{\exp(\text{Score}(x, d))}{\sum\limits_{d' \in D} \exp(\text{Score}(x, d'))}
\end{equation}

\vspace{1em}
\noindent The full process is summarized in Algorithm~\ref{alg:rag}.

\begin{algorithm}[H]
\caption{Retrieval-Augmented Generation (RAG) Pipeline}
\label{alg:rag}
\begin{algorithmic}[1]
\Require Input query $x$, corpus $\mathcal{C}$, embedding model $\phi$, retriever $R$, generator $\mathcal{G}$, top-$k$ parameter $k$
\Ensure Generated response $\hat{y}$
\Statex
\State // \textbf{Offline: Index Construction}
\ForAll{$d_i \in \mathcal{C}$}
    \State $v_i \gets \phi_d(d_i)$ \Comment{Encode documents}
\EndFor
\State Store $\{v_i\}$ in vector index $\mathcal{I}$

\Statex
\State // \textbf{Online: Inference Pipeline}
\State $q \gets \phi_q(x)$ \Comment{Encode query}
\State $D \gets R(q, \mathcal{I})$ \Comment{Retrieve top-$k$ docs}
\State $C \gets \text{concat}(x, D)$ \Comment{Combine query and retrieved context}
\State $\hat{y} \gets \mathcal{G}(C)$ \Comment{Generate grounded output}
\State \Return $\hat{y}$
\end{algorithmic}
\end{algorithm}

This retrieval-augmented framework enables the system to dynamically adapt to new information in real-time, bypassing the need to retrain the parameters $\theta$, and thereby improving factual grounding, explainability, and domain adaptability.

\subsection{Historical Development}

The conceptual foundations of RAG can be traced back to early memory-based AI systems, where knowledge was encoded directly into model parameters. While effective for many language tasks, these models suffered from inherent rigidity - once trained, they could not incorporate new knowledge without undergoing extensive retraining. This limitation spurred interest in external memory architectures, where retrieval mechanisms could extend the knowledge base without altering model weights.

Initial retrieval systems relied on sparse, term-matching algorithms such as TF-IDF and BM25, widely adopted during the 1990s and 2000s. These methods enabled a  large-scale document search based on lexical similarity and laid the groundwork for information retrieval in natural language tasks. However, their reliance on surface-level matching limited their ability to handle semantically rich or paraphrased inputs.

\begin{figure*}[t] 
    \centering
    \includegraphics[width=0.8\textwidth]{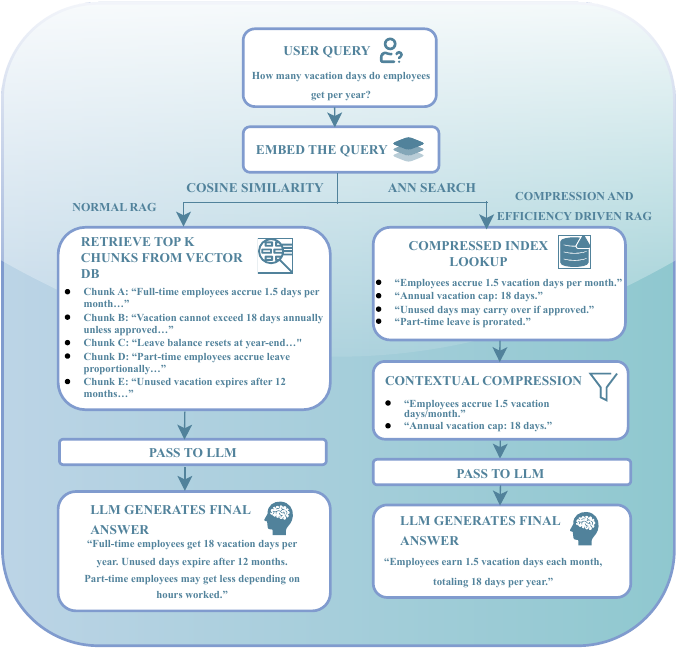} 
    \caption{Schematic of Compression and Efficiency-Driven RAG systems.}
    \label{fig:compression} 
\end{figure*}

The introduction of neural retrieval models in the late 2010s marked a transformative step. Dense retrieval systems such as DPR employed deep neural networks to map queries and documents into a shared embedding space, allowing for semantic similarity computation via vector distance. This innovation significantly improved retrieval quality and enabled more effective integration with downstream language generation tasks.

The RAG framework was formally introduced in 2020 by researchers at Facebook AI and OpenAI, integrating dense retrieval and generative LLMs into a unified architecture. Unlike earlier pipeline-based approaches, RAG treated retrieval and generation as tightly coupled stages within a single inference workflow. This advancement allowed models to generate responses that were not only contextually coherent but also grounded in verifiable external knowledge, leading to improved performance on knowledge-intensive benchmarks.

Recent advancements have focused on expanding the scope of retrieval, including long-context retrieval systems capable of processing entire documents or multiple knowledge sources. These developments are especially relevant for applications requiring complex reasoning, multi-step inference, and real-time adaptation. As the field continues to evolve, future research is expected to emphasize retrieval efficiency, fairness and bias mitigation, and deeper integration of RAG systems into practical deployments requiring transparency, trust, and explainability.

\section{Taxonomy of RAG-LLM Architectures}
\label{sec:taxonomy}

\subsection{Compression and Efficiency-Driven RAG}
\label{sec:compression-efficiency}

As RAG architectures scale to handle increasingly large knowledge corpora, one of the central challenges becomes maintaining retrieval efficiency without compromising response quality. This taxonomy category encompasses methods that reduce computational overhead, minimize retrieval latency, and optimize the selection of relevant context. Compression- and efficiency-driven RAG systems focus on refining how, what, and how much to retrieve - balancing performance with scalability in real-world applications.

We define a retrieval and compression-aware objective for Compression and Efficiency-Driven RAG systems as follows:
\begin{equation}
\min_{\theta, \phi} \; \mathbb{E}_{q \sim \mathcal{Q}} \left[ \sum_{i=1}^{k} \mathcal{U}(q, d_i) \cdot \log \left( \frac{p_\theta(d_i \mid q)}{C_\phi(d_i)} \right) \right]
\end{equation}

\noindent where:
\begin{itemize}
  \item \( q \in \mathcal{Q} \): input query drawn from the query distribution.
  \item \( d_i \): the \(i\)-th retrieved document for query \(q\).
  \item \( p_\theta(d_i \mid q) \): retrieval probability of document \(d_i\) given query \(q\), parameterized by retriever parameters \( \theta \).
  \item \( \mathcal{U}(q, d_i) \in [0,1] \): utility function measuring the relevance or helpfulness of \(d_i\) for query \(q\).
  \item \( C_\phi(d_i) > 0 \): compression cost or token-level cost of including document \(d_i\), parameterized by compression module \( \phi \).
  \item \( \log \left( \frac{p_\theta(d_i \mid q)}{C_\phi(d_i)} \right) \): log-efficiency score balancing relevance and cost.
\end{itemize}

This objective prioritizes high-utility, low-cost documents for inclusion in the context window, reflecting the core trade-offs in Compression and Efficiency-Driven RAG.

\vspace{1em}
\noindent The overall pipeline for compression-aware retrieval is formalized in Algorithm~\ref{alg:compression-rag}.

\begin{algorithm}[H]
\caption{Compression-Aware Retrieval-Augmented Generation}
\label{alg:compression-rag}
\begin{algorithmic}[1]
\Require Query $q$, corpus $\mathcal{C}$, retriever parameters $\theta$, compressor parameters $\phi$, top-$k$ retrieval size
\Ensure Compressed document context $D_{\text{comp}}$ for generation
\State $q_{\text{vec}} \gets \phi_q(q)$ \Comment{Encode query}
\State $D \gets \text{RetrieveTopK}(q_{\text{vec}}, \mathcal{C}, \theta)$
\ForAll{$d_i \in D$}
    \State Compute $\mathcal{U}(q, d_i)$ \Comment{Utility or relevance score}
    \State Compute $C_\phi(d_i)$ \Comment{Compression or cost estimate}
    \State Compute efficiency score $E_i \gets \log\left( \frac{p_\theta(d_i \mid q)}{C_\phi(d_i)} \right)$
\EndFor
\State $D_{\text{comp}} \gets \text{SelectTopDocuments}(D, \text{based on } E_i)$
\State \Return $D_{\text{comp}}$
\end{algorithmic}
\end{algorithm}

Algorithm~\ref{alg:compression-rag} captures the key operations of Compression and Efficiency-Driven RAG. The process begins by encoding the query and retrieving a candidate set of $k$ documents from a large corpus. For each candidate, a utility function assesses its semantic relevance to the query, while a compression model estimates the cost of including that document in the LLM's input context (e.g., token length, redundancy, entropy). The efficiency score, defined as the log-ratio of retrieval probability to compression cost, quantifies the relevance-to-cost trade-off. The final step involves selecting the subset of documents with the highest efficiency scores - i.e., documents that offer the best informativeness per unit of computational or memory cost. This approach enables scalable generation under limited context budgets, making RAG viable for latency-sensitive or resource-constrained settings.

Early retrieval-based methods that contributed to this class often preceded the formalization of RAG but laid essential groundwork through implicit alignment with its core ideas. A notable contribution was REINA (Retrieving from the traINing datA), which enhanced NLP task performance by retrieving semantically similar labeled examples from the training set \cite{reina2022}. Though it avoided external corpora, its focus on instance-level reuse of training data resonated with themes found in works such as kNN-LM \cite{khandelwal2020knn}, which similarly emphasized local adaptation through retrieval. Both methods exposed the computational bottlenecks involved in large-scale index traversal - a constraint that later frameworks like REALM \cite{guu2020realm} and KILT \cite{petroni2021kilt} explicitly addressed in scalable RAG contexts.

Expanding beyond dataset-internal retrieval, large-scale retrieval emerged as a central strategy in studies like {Improving Language Models by Retrieving from Trillions of Tokens} \cite{retrieval-trillions2022} and concurrent work by Borgeaud et al. on GopherCite \cite{borgeaud2022improving}, which showed the advantages of augmenting language models with web-scale corpora. These approaches demonstrated significant improvements in domain-specific and low-resource tasks, echoing earlier themes in memory-augmented networks \cite{weston2014memory} and retrieval-based QA systems such as DrQA \cite{chen2017reading}. Although they did not use the RAG terminology explicitly, these efforts collectively marked a transition toward scalable external retrieval - a principle that underpins many web-scale RAG frameworks today.

Handling retrieval across heterogeneous sources was explored in {RAG across Heterogeneous Knowledge} \cite{yu-2022-retrieval}, where multi-source integration mechanisms were developed to cope with diverse formats, noise levels, and domain-specific gaps. This thread aligns with earlier work on knowledge integration from multiple corpora \cite{lee2019latent, izacard2021distilling}, which also sought to balance retrieval fidelity and generation quality. The problem of heterogeneity has since become central in hybrid systems like UnifiedQA \cite{khashabi2020unifiedqa} and GTR \cite{ni2021large}, positioning retrieval diversity as both a challenge and a potential enabler for factual depth and task robustness in RAG systems.

To reduce the annotation burden in supervised retrieval pipelines, PROMPTAGATOR \cite{dai2022promptagator} introduced a few-shot dense retrieval pipeline that leveraged LLM prompting to bootstrap query generation with as few as eight labeled examples. This strategy intersects with ideas from GENRE \cite{de2020autoregressive} and InstructGPT \cite{ouyang2022training}, which similarly demonstrated how instruction tuning and label-free methods could drive performance in data-scarce scenarios. Collectively, such efforts highlight the growing trend of lowering supervision costs through LLM-driven heuristics - critical for scalable RAG deployment across both high- and low-resource domains.

Complementary insights on retrieval fidelity were provided by research investigating retrieval noise. {LLMs Can Be Easily Distracted by Irrelevant Context} \cite{shi2023llmdistraction} illustrated the risks of unfiltered retrieval, echoing earlier concerns raised in the context of distractor-based QA datasets \cite{rajpurkar2018know} and adversarial contexts \cite{jia2017adversarial}. These studies collectively emphasize the importance of query refinement, selective attention, and context filtering - techniques that are increasingly incorporated into modern RAG systems, such as through learned retriever scoring \cite{izacard2022feww} and re-ranking mechanisms \cite{maillard2021multi}.

Collectively, these contributions established the defining characteristics of compression- and efficiency-driven RAG: minimal reliance on extensive supervision, scalable retrieval over heterogeneous and massive corpora, and resilience to noisy or misleading contexts. The transition from tightly curated datasets to open-domain, multi-source settings was further bolstered by improvements in retriever architecture (e.g., ColBERT \cite{khattab2020colbb}) and retrieval latency management (e.g., ATLAS \cite{izacard2022few}). These techniques laid the groundwork for high-throughput RAG pipelines capable of sustaining generation quality while optimizing computational cost.

Building on the foundational developments of 2022, subsequent research in 2023 advanced the design of efficiency- and compression-oriented RAG systems through the incorporation of domain specialization \cite{asai2023retrieval}, tighter retriever-generator coupling via shared latent representations \cite{busta2023learning}, and improved generalization through few-shot and unsupervised learning \cite{menick2022teaching}. These innovations reflect a deliberate shift from generic retrieval paradigms to task-aware, context-sensitive architectures, informed by broader advances in prompt engineering \cite{zhou2022least}, curriculum learning \cite{xu2023mixed}, and contrastive learning-based retrievers - further underscoring the evolving sophistication of the RAG landscape.

One important direction focused on applying RAG in low-data environments. The integration of retrieval with few-shot learning was explored in the context of model synthesis for domain-specific languages (DSLs), where annotated data is often scarce \cite{dsl1}. By retrieving in-domain exemplars to guide generation, these systems tackled the challenge of out-of-distribution generalization - a recurring issue in deploying LLMs for specialized tasks. Retrieval thus acted as a mechanism for implicit supervision, improving generation quality in underrepresented domains \cite{atlas, retro}.

In parallel, structural optimization of LLMs gained prominence as a means to reduce computational overhead in RAG systems. Structured pruning methods were proposed to compress generative backbones without significant loss in performance \cite{pruning2023, retroprun}, offering a lightweight alternative for latency-sensitive applications. Complementary to these efforts, augmentation-aware retrievers were introduced to improve the alignment between retrieved context and model generation objectives \cite{retriever2023, adaptive-retrieval, dpr}. These approaches emphasized the importance of treating the retriever and generator as co-adaptive modules rather than disjoint components, a shift also explored in end-to-end trained frameworks such as REALM and DensePhrases \cite{realm, densephrases}.

Researchers also extended RAG principles to tasks beyond traditional knowledge-intensive domains. Prompt-guided retrieval strategies were proposed to enhance performance in non-knowledge-intensive tasks such as classification and generation, by using prompt reranking or retrieval-augmented prompting to better align the retrieved context with task-specific requirements \cite{prompt-guided2023, promptagator, reranker-rag}. This extension illustrated that retrieval can improve not only factual grounding but also task relevance, controllability, and input conditioning in domains with less structured supervision.

To handle structured information, models like SANTA were introduced to pretrain language models with an inductive bias toward structured data formats, thereby improving dense retrieval from semi-structured corpora such as tables or knowledge graphs \cite{santa20231, tabert, kg-rag}. These techniques expanded the applicability of RAG systems beyond unstructured text corpora and into more structured, domain-specific repositories such as scientific tables, product catalogs, and relational databases.

Later work further emphasized generalization and open-domain adaptability. UPRISE introduced a universal prompt retrieval system for zero-shot evaluation tasks, leveraging retrieval to design effective prompts without task-specific tuning \cite{uprise20231, unify, metaprompt}. In the recommendation domain, RAG techniques were adapted for open-world personalization by incorporating retrieval-augmented knowledge into recommender pipelines \cite{rag-recsys20231, rag4rec, recgpt}, demonstrating the versatility of RAG across user-centric applications and longitudinal interaction modeling.

The trajectory continued in 2024 with an increasing emphasis on retrieval granularity, efficiency at scale, and domain-aware evaluation. A central thread across the year was the refinement of query construction and retrieval mechanics to improve model adaptability and contextual relevance \cite{rag-survey2024, adaptiveq, retriever-tuning}.

New architectures such as RAPTOR introduced hierarchical retrieval mechanisms that structured document access as recursive tree-based abstractions \cite{raptor2024}, enabling more efficient reasoning over multi-level content. Complementing this, RQ-RAG proposed query refinement modules that dynamically rewrote retrieval queries to improve document relevance and retrieval quality \cite{rqrag20241, rerag, dynamicretrieval}. These developments addressed longstanding limitations around shallow retrieval, static queries, and brittle document ranking.

Retrieval robustness across datasets was further explored in studies evaluating context selection strategies across multiple domains \cite{retrieval-strategies20241, retrieval-ablation, selection-bench}. This line of work emphasized that a one-size-fits-all retrieval policy fails to generalize and called for adaptive retrieval mechanisms informed by dataset characteristics, task difficulty, and user intent alignment.

Structured and domain-specific retrieval gained momentum as RAG applications expanded into finance, legal, and multilingual domains. Chunking strategies tailored to long-form financial disclosures were introduced to enhance token-level granularity and improve evidence localization in dense corpora \cite{financial-rag20241, longrag, lexrag}. Benchmarks such as LegalBench-RAG and Arabic-specific evaluations highlighted the limitations of current retrievers when faced with highly structured, non-English legal documents \cite{arabicrag20241, legalbench2024}. These works reinforced the importance of structure-sensitive retrieval and cross-lingual alignment, especially in regulatory and compliance-heavy use cases.

A parallel research axis emphasized tooling and rapid experimentation. Modular frameworks such as FlashRAG, RAG Foundry, and RAGLAB emerged to support configurable pipelines with interchangeable retrievers, encoders, and generation heads \cite{flashrag2024, ragfoundry2024, raglab2024}. These toolkits, inspired by prior efforts like LangChain and Haystack \cite{haystack, langchain}, significantly lowered the barrier to entry for applied RAG, enabling reproducible baselines and faster iteration cycles for both academic and industrial deployments.

Efficiency innovations included novel compression and optimization techniques designed to preserve utility while reducing token load and inference cost. xRAG introduced extreme context distillation, encoding high-value evidence into ultra-compact representations without compromising downstream generation fidelity \cite{xrag2024, lite-rag}. In parallel, superposition prompting explored the use of parallelized retrieval heads with confidence-aware aggregation to reduce latency and document redundancy \cite{ multiquery}. Stochastic RAG introduced an information-theoretic utility maximization objective, selectively retrieving documents based on expected generation impact \cite{retriever-reward}.

Alongside these modeling advances, 2024 witnessed the maturation of evaluation frameworks dedicated to retrieval-augmented systems. RAGAS proposed automated, reference-free metrics that jointly assess retrieval quality and answer generation fidelity, offering scalable alternatives to human annotation \cite{ragas2024, faithfulness-metrics}. RAGBench introduced a multi-domain suite for explainable, end-to-end RAG evaluation spanning science, law, and open-domain QA \cite{ragbench2024, retrievalqa}. Additional efforts underscored the pivotal role of embedding model selection in retrieval relevance, especially when deploying across domains with distinct semantics \cite{embedding-eval2024, embeddiff}.

These advances paved the way for novel use cases beyond traditional QA. Retrieval-augmented test generation was explored in software engineering and fuzzing pipelines to improve code coverage and fault detection \cite{testgen2024, codeprompt}, while generative retrieval was deployed in recommender systems to synthesize user-aligned candidate items from latent memory banks \cite{genrev2024, rag4rec}. Synthetic dataset generation using retrieval-guided sampling also gained traction as a scalable means to bootstrap low-resource domains \cite{synthesizrr2024, datafab}.

Across 2023 and 2024, the RAG landscape matured into a multi-faceted ecosystem. Compression-aware architectures, query-refinement modules, structure-sensitive retrieval techniques, modular toolkits, and domain-general evaluation benchmarks collectively transformed RAG from a generic pipeline into a versatile, scalable, and efficient augmentation paradigm. These developments cement RAG’s position not only as a retrieval framework but as a foundational taxonomy for knowledge-grounded language generation.

\subsection{Defensive RAG: Addressing Bias, Vulnerabilities, and Privacy}
\label{sec:defensive-rag}

\begin{figure*}[t]
    \centering
    \includegraphics[width=0.9\textwidth]{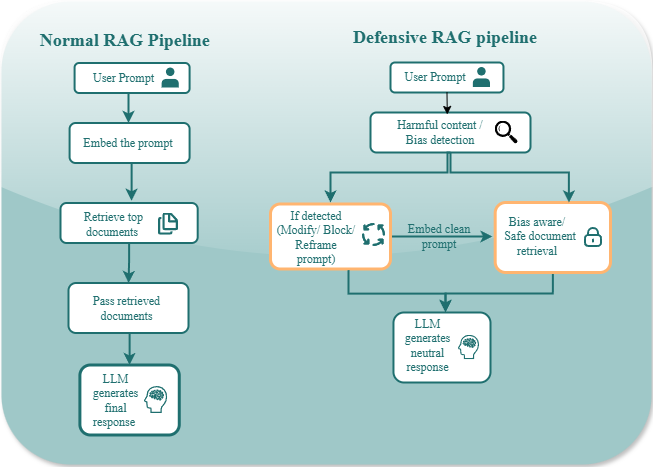}
    \caption{Comparison between standard and defensive RAG pipelines, illustrating safeguards for harmful content detection, prompt sanitization, and bias-aware document retrieval.}
    \label{fig:defense}
\end{figure*}

As RAG systems extend into real-world applications, the need to address issues of factual reliability, contextual vulnerability, and ethical integrity has given rise to a new class of architectures we term {Defensive RAG}. This taxonomy encompasses approaches that mitigate hallucinations, combat misinformation, filter harmful content, and address fairness and privacy concerns. Defensive RAG systems aim to make RAG more robust, trustworthy, and socially responsible.

We define a bias-mitigation, vulnerability-aware, and privacy-preserving objective for Defensive RAG systems as follows:

\begin{align}
\mathcal{L}_{\text{DefRAG}} 
&= \mathbb{E}_{q \sim \mathcal{Q}} \Bigg[
    \sum_{d \in \mathcal{D}_q} 
    \Big( (1 - \mathcal{T}(d)) 
    + \mathcal{P}(d) 
    + \mathcal{B}(d) \Big)
    \cdot R(q, d)
\Bigg]
\end{align}

\noindent where:
\begin{itemize}
  \item $\mathcal{Q}$ is the distribution over user queries $q$.
  \item $\mathcal{D}_q$ is the set of retrieved documents for query $q$.
  \item $R(q, d) \in [0, 1]$ is the retrieval relevance score between query $q$ and document $d$.
  \item $\mathcal{T}(d) \in [0, 1]$ is the probability that document $d$ is \textbf{non-toxic}.
  \item $\mathcal{P}(d) \in [0, 1]$ is the probability that document $d$ contains \textbf{private or sensitive content}.
  \item $\mathcal{B}(d) \in [0, 1]$ is a bias score quantifying demographic or cultural bias in $d$.
\end{itemize}

\noindent This loss function penalizes the inclusion of toxic, private, or biased documents - even if they are topically relevant - thereby promoting safe, fair, and responsible information retrieval in RAG pipelines.

\vspace{1em}
\noindent Algorithm~\ref{alg:defensive-rag} outlines the core flow of a defensive RAG retrieval module that prioritizes trustworthiness over raw topical relevance.

\begin{algorithm}[H]
\caption{Bias-, Privacy-, and Safety-Aware Defensive RAG}
\label{alg:defensive-rag}
\begin{algorithmic}[1]
\Require Query $q$, corpus $\mathcal{C}$, scoring models $\mathcal{T}, \mathcal{P}, \mathcal{B}$, relevance function $R$
\Ensure Filtered document set $D_{\text{safe}}$, generated response $\hat{y}$
\State $q_{\text{vec}} \gets \phi_q(q)$ \Comment{Encode query}
\State $D_{\text{cand}} \gets \text{RetrieveTopK}(q_{\text{vec}}, \mathcal{C})$
\ForAll{$d \in D_{\text{cand}}$}
    \State Compute relevance $R(q, d)$
    \State Compute risk score:
    $S_d \gets (1 - \mathcal{T}(d)) + \mathcal{P}(d) + \mathcal{B}(d)$
    \State Compute safe-weighted score:
    $\tilde{R}(q, d) \gets R(q, d) \cdot (1 - S_d)$
\EndFor
\State $D_{\text{safe}} \gets \text{SelectTop}(D_{\text{cand}}, \tilde{R})$
\State $\hat{y} \gets \mathcal{G}(q, D_{\text{safe}})$
\State \Return $\hat{y}$
\end{algorithmic}
\end{algorithm}

Algorithm~\ref{alg:defensive-rag} implements a filtered document selection loop to enforce ethical and safety constraints during retrieval. After retrieving a set of top-$k$ candidates based on standard relevance, each document is evaluated using three classifiers: toxicity ($\mathcal{T}$), privacy leakage ($\mathcal{P}$), and demographic/cultural bias ($\mathcal{B}$). A composite risk score is calculated and used to reweight each document’s relevance, penalizing those that are likely to introduce harm. The safest, most relevant documents are selected for final generation. This defensive formulation is particularly critical in domains such as healthcare, education, and public policy, where unfiltered retrieval may lead to hallucinations, misinformation, or ethical breaches.

As RAG systems gained traction, a parallel line of inquiry emerged around defensive mechanisms, seeking to mitigate risks such as hallucinations, contextual misinformation, bias amplification, and privacy leakage. Early foundational work emphasized the role of structured context retrieval in enhancing zero-shot generalization. UPRISE introduced universal prompt retrieval as a mechanism to dynamically select task-relevant prompts from curated pools, enabling LLMs to generalize to novel tasks without handcrafted prompts \cite{uprise2022}. While not document-level RAG per se, this prompt-centric retrieval framework anticipated more advanced pipelines where retrieval is used not merely for grounding but also for guiding model behavior under uncertain or unstructured conditions.

Attention soon shifted toward the vulnerabilities introduced by noisy or adversarial contexts. Studies demonstrated that LLMs exhibit high sensitivity to irrelevant, misleading, or conflicting retrievals, which can degrade factual fidelity. To counter this, frameworks like FILCO proposed fine-grained post-retrieval filtering modules to select documents based on factual alignment and coherence metrics \cite{filco2023}. Complementing this, HaluEval benchmarked hallucination frequency across diverse LLMs and RAG variants, underscoring that even minor perturbations in context can induce significant generation drift \cite{halueval2023, truthscore}. These insights formed the basis for filtering-aware retrieval modules and context sanitization pipelines now considered standard in high-stakes deployments.

A related thread explored how LLMs themselves can act as retrieval-aware agents, not just generators. Research revealed that while few-shot extraction and reasoning tasks remain challenging for LLMs, their effectiveness as semantic rerankers is markedly stronger in noisy retrieval environments \cite{reranker2023, llm4retrieval}. This gave rise to hybrid architectures where candidate passages are first retrieved via dense or sparse indices, then reranked or rescored by the LLM before final generation. Frameworks like SELF-RAG integrated this loop into a self-reflective cycle, combining retrieval, generation, and critique modules that iteratively refine outputs through internal feedback mechanisms \cite{selfrag2023, reflexion2023}. These designs showcased the potential of retrieval-generation co-adaptation, particularly for increasing factuality and robustness.

Beyond performance, an expanding body of work has interrogated fairness, bias, and ethical integrity in retrieval-augmented generation. Studies revealed systemic biases in both retrieved content and generated responses - ranging from cultural erasure to demographic stereotyping. Works such as BiasBench and MuslimBias highlighted the asymmetric treatment of sociocultural groups in RAG outputs, even when retrieval mechanisms were ostensibly neutral \cite{bias-chatgpt2023, muslim-bias2023}. Parallel critiques of evaluation pipelines, including those driven by LLM-based scoring, questioned their objectivity and reproducibility, especially in domains involving sensitive group identities \cite{llms-not-fair2023, redteaming-rag}. These studies urged for ethics-aware retrieval filters, transparency in scoring protocols, and greater scrutiny in benchmark design.

Privacy and safety also emerged as critical concerns in domains like healthcare, law, and defense. In such settings, poorly controlled retrieval can surface confidential, outdated, or misleading content, leading to downstream ethical and legal liabilities. Calls for domain-specific retrieval governance, including knowledge base curation and trust-weighted ranking algorithms, have grown louder in recent literature \cite{llms-not-fair2023, domain-safety-rag}. These efforts align with a broader vision of Defensive RAG, where retrieval is not just a tool for grounding but a gatekeeper of factual integrity, safety, and societal responsibility.
\begin{figure*}[t] 
    \centering
    \includegraphics[width=0.9\textwidth]{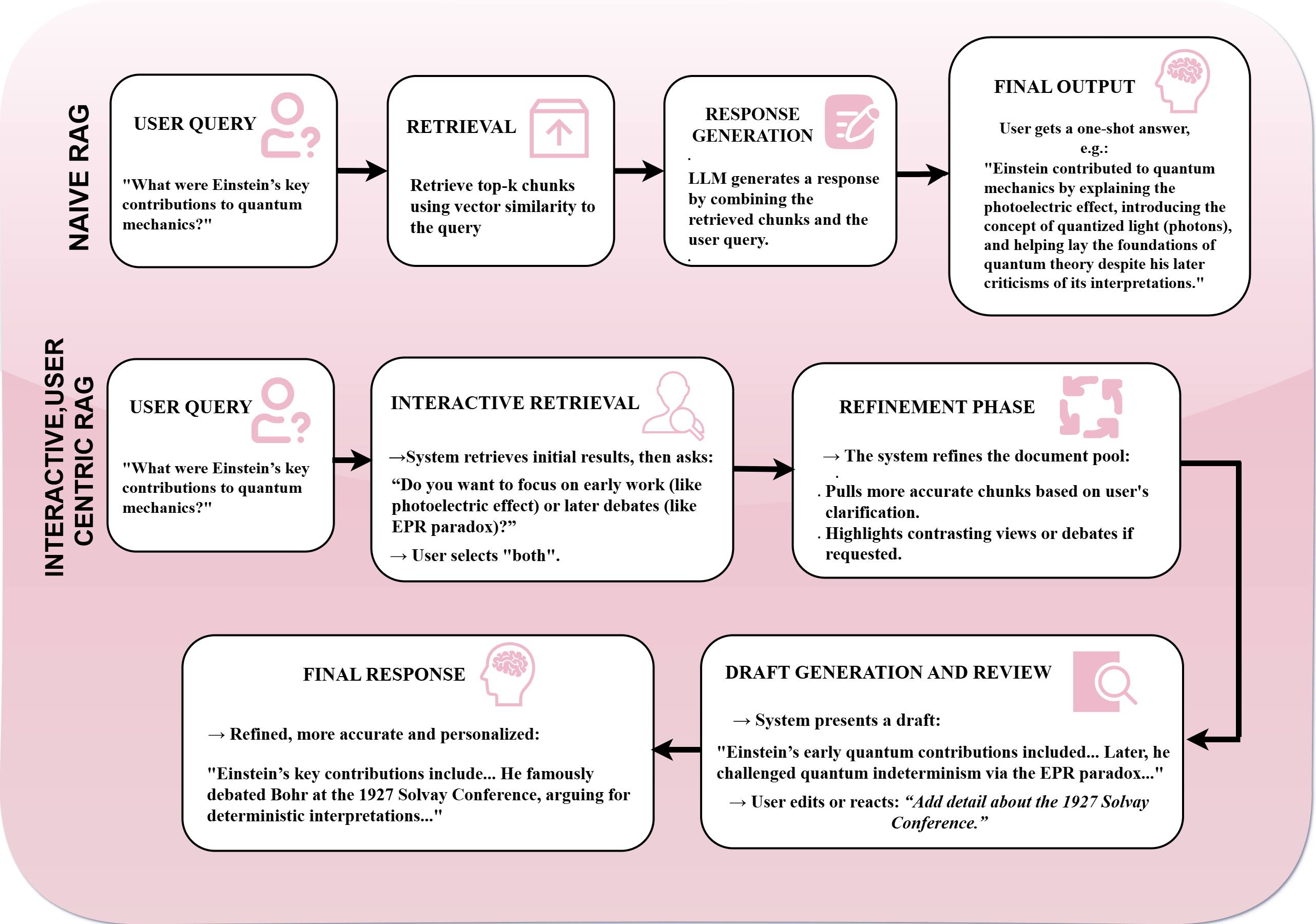} 
    \caption{Comparison between Naive RAG and Interactive, User-Centric RAG architectures. While Naive RAG retrieves top-k chunks via vector similarity and generates static responses, the Interactive RAG framework introduces a dynamic retrieval loop. It incorporates user intent clarification, refinement of source documents based on feedback, and iterative draft-review cycles. This enables more accurate, personalized, and context-aware answers, highlighting the value of user-in-the-loop mechanisms for retrieval-augmented generation systems.}
    \label{fig:interactive} 
\end{figure*}

To address emerging vulnerabilities in retrieval-augmented generation, a wave of research introduced self-aware retrieval mechanisms, where models estimate their own uncertainty and selectively trigger retrieval when needed. This culminated in the formalization of self-knowledge-guided retrieval, a framework where introspective signals inform when and how to retrieve external evidence \cite{selfknowledge2023, uncertainty-routing}. These systems mark a shift from passive retrieval to retrieval-as-needed, aligning retrieval activation with confidence thresholds.

In 2024, Defensive RAG matured across dimensions such as privacy, robustness, preference alignment, and systematic error tracking. As retrieval assumed a central role in LLM pipelines, researchers identified new failure modes spanning contextual noise, security gaps, and fairness drift, reinforcing the need for retrieval-aware safety layers \cite{defrag2024, saferag}.

Comparative studies investigated the tradeoffs between fine-tuning and retrieval-based augmentation for integrating external knowledge. Results revealed nuanced behavior, with retrieval outperforming fine-tuning on sparse domains while underperforming on deeply specialized ones \cite{ft-retrieval-benchmark}. In parallel, practical bottlenecks - such as hallucination persistence, unclear attribution, and retrieval irrelevance - were codified in taxonomies of known RAG failure points, prompting more defensive engineering patterns \cite{seven-failure-points2024}.

Privacy and information leakage became key areas of concern. Researchers demonstrated how RAG systems may inadvertently surface sensitive or misaligned information through unfiltered document access or compromised retrieval sources \cite{privacy-rag2024, ragleak}. To counteract this, methods such as synthetic retrieval corpora \cite{synthetic-rag2024}, access control-aware retrievers \cite{access-control-rag2024}, and adversarial input sanitization were proposed to guard against data poisoning and extraction threats \cite{badrag2024, poisonedrag2024}.

Defensive strategies were also applied to the handling of noisy or low-salience retrievals. Traditional filtering models suppressed irrelevant documents, but surprisingly, controlled noise injection - adding neutral distractors - was found to help calibrate LLM uncertainty and improve final output robustness \cite{noise-rag2024, distractor-blend}. These findings challenged prior assumptions about retrieval purity, suggesting that selective stochasticity could enhance generative resilience.

In pursuit of deeper alignment, preference-guided retrieval architectures emerged. These models ranked candidates not solely by surface relevance but also by compatibility with LLM inference patterns, mitigating the risk of misleading but on-topic content dominating the context window \cite{dual-preference-rag2024, alignment-ranker}. Concurrently, studies revealed new retrieval denial-of-service attacks, where distractor documents were inserted to degrade retrieval quality or overwhelm rerankers \cite{jam-rag2024}, underscoring the necessity for robust scoring and defense mechanisms.

Finally, verification-driven pipelines were proposed to strengthen factual grounding. Chain-of-verification techniques introduced intermediate validation stages where claims retrieved from external sources were fact-checked prior to generation \cite{chain-verification2024, retrieval-checker}. These techniques extended the principles of SELF-RAG and were expanded to domains such as multilingual consistency \cite{multilingual-awareness2024} and factual boundary modeling, where models learn to estimate and report uncertainty thresholds for claims with ambiguous evidence \cite{factual-boundaries2024}.

Several benchmarking efforts emerged to ensure that retrieval quality, factual consistency, and robustness are properly evaluated. IRSC and UDA proposed real-world and zero-shot benchmarks for measuring RAG resilience \cite{irsc2024, uda2024}, while BERGEN provided a modular library for unified evaluation \cite{bergen2024}. Domain-specific metrics, such as Face4RAG for Chinese factuality evaluation \cite{face4rag2024}, demonstrated growing interest in cross-lingual and culturally sensitive RAG validation.

Finally, ethical and fairness concerns were increasingly studied. The role of retrieval ranking in reinforcing information exposure bias was examined in {Towards Fair RAG}, which highlighted how unfair ranking algorithms can skew generated content distributions \cite{fairrag2024}. This work underscored the importance of incorporating fairness-aware objectives into both retrieval scoring and generation pipelines.

Collectively, these developments define the emerging field of Defensive RAG. By focusing on filtering, reranking, uncertainty modeling, bias mitigation, and ethical safeguards, these systems move beyond naive retrieval augmentation to more principled, context-sensitive, and risk-aware architectures. From query-level preference alignment and privacy-aware retrieval to adversarial resilience and fairness benchmarking, Defensive RAG evolved into a fully-fledged design space. These efforts make RAG more trustworthy and pave the way for safe deployment in high-risk domains like healthcare, finance, and law.

\subsection{Interactive, User-Centric RAG Systems}
\label{sec:interactive-user-centric}

Traditional RAG systems are typically designed around static retrieval pipelines that optimize document relevance with respect to a query. However, such pipelines often ignore the dynamic, user-driven nature of real-world interactions - where user goals shift, context evolves, and retrieved knowledge must adapt accordingly. This taxonomy captures RAG architectures that are designed to be interactive, user-aware, and context-sensitive, enabling LLMs to refine, adapt, or personalize their retrieval and generation processes based on user intent and evolving input.

\begin{align}
\mathcal{L}_{\text{InterRAG}} 
&= \mathbb{E}_{(q, u) \sim \mathcal{U}} \Bigg[
    \sum_{d \in \mathcal{D}_{q,u}} 
    \Big( \alpha \cdot S(q, d)
\notag\\
&\qquad\qquad
    + \beta \cdot H(u, d)
    + \gamma \cdot F(d, u) \Big)
    \cdot \pi_{u}(d \mid q)
\Bigg]
\end{align}

\noindent where:
\begin{itemize}
  \item $\mathcal{U}$ is the distribution over query-user pairs $(q, u)$.
  \item $\mathcal{D}_{q,u}$ is the set of retrieved documents for query $q$ and user $u$.
  \item $S(q, d) \in [0, 1]$ is the semantic relevance score between query $q$ and document $d$.
  \item $H(u, d) \in [0, 1]$ is the historical interaction alignment score between user $u$ and document $d$.
  \item $F(d, u) \in [0, 1]$ is the feedback compatibility score estimating how well document $d$ aligns with user preferences.
  \item $\pi_{u}(d \mid q)$ is the user-conditioned document selection policy.
  \item $\alpha, \beta, \gamma \in \mathbb{R}_{+}$ are tunable weights controlling the trade-off between query relevance, historical consistency, and personalization.
\end{itemize}

\noindent This objective function captures the interactive RAG goal of aligning document selection with evolving user intent, interaction history, and personalization signals.

\vspace{1em}
\noindent The end-to-end interactive personalization pipeline is described in Algorithm~\ref{alg:interactive-rag}.

\begin{algorithm}[H]
\caption{Interactive, User-Aware RAG Pipeline}
\label{alg:interactive-rag}
\begin{algorithmic}[1]
\Require Query $q$, user profile $u$, document corpus $\mathcal{C}$, selection policy $\pi_u$, scoring functions $S$, $H$, $F$
\Ensure Personalized context $D_{q,u}$ and generated response $\hat{y}$
\State $q_{\text{vec}} \gets \phi_q(q)$ \Comment{Encode query}
\State $D_{\text{cand}} \gets \text{RetrieveTopK}(q_{\text{vec}}, \mathcal{C})$
\ForAll{$d \in D_{\text{cand}}$}
    \State Compute $S(q, d)$ \Comment{Query-document semantic relevance}
    \State Compute $H(u, d)$ \Comment{User history alignment score}
    \State Compute $F(d, u)$ \Comment{Feedback/personalization score}
    \State Compute total score:
    $R_d \gets \alpha S(q, d) + \beta H(u, d) + \gamma F(d, u)$
\EndFor
\State $\pi_u(d \mid q) \gets \text{Softmax over } R_d$
\State Sample $D_{q,u} \sim \pi_u(d \mid q)$ \Comment{Personalized document selection}
\State $\hat{y} \gets \mathcal{G}(q, D_{q,u})$ \Comment{Generate response with selected context}
\State \Return $\hat{y}$
\end{algorithmic}
\end{algorithm}

Algorithm~\ref{alg:interactive-rag} outlines a retrieval-generation loop that integrates user feedback, profile information, and intent modeling to personalize outputs. Given a query and user identity, candidate documents are scored not just by semantic relevance to the query ($S(q,d)$), but also by how well they align with the user’s historical interactions ($H(u,d)$) and expressed or inferred preferences ($F(d,u)$). A learned or heuristic document selection policy $\pi_u$ combines these scores and samples a final set of documents $D_{q,u}$ to serve as input context for the generator. This interactive loop enables adaptation over sessions, dynamic query expansion, and alignment with user expectations - making it well-suited for recommendation, conversational agents, and domain-specific assistants.

Early investigations into prompt-based language modeling revealed limitations in conventional few-shot prompting techniques, particularly their rigidity and dependency on hand-crafted inputs. The emergence of metaprompting frameworks, as exemplified in Prompt Programming for LLMs: Beyond the Few-Shot Paradigm, marked a shift toward model-driven instruction generation, where LLMs construct task-specific prompts aligned with user objectives \cite{metaprompt2021}. This strategy demonstrated strong zero-shot generalization capabilities, rivaling few-shot baselines across reasoning tasks \cite{replug2023, promptsource2022}. However, despite enhancing the alignment between model behavior and user goals, these techniques were fundamentally reactive, confined to static instruction templates that offered limited capacity for information-seeking or clarification under uncertainty \cite{reclor2020, promptengineering2021}.

To address this passivity, ISEEQ proposed an interactive retrieval paradigm where LLMs actively formulate Information-Seeking Questions (ISQs) to augment under-specified queries \cite{iseeq2021}. By leveraging structured knowledge graphs and meta-retrieval signals, the model dynamically clarified its own generation space, offering coherent, relevance-driven questions that improved downstream accuracy. This methodology paralleled developments in multi-hop QA \cite{rag2020, baleen2022}, where intermediate questions guide retrieval, and generative feedback loops enhance contextual precision \cite{selfask2021}. ISEEQ further integrated adversarial reinforcement learning and retrieval-aware loss functions to optimize ISQ quality, facilitating transitions from static prompting to interactive, multi-turn query resolution frameworks \cite{rarr2022, activeprompting2023}. This represented a conceptual realignment: from instruction execution to dialogue-centric generation pipelines, where the model initiates external knowledge access based on internal ambiguity.

Subsequent research extended interactive RAG architectures into multimodal environments, introducing real-time systems capable of ingesting and synchronizing diverse signal types. One such milestone was Lingua, a speech-centric RAG system designed for end-to-end STS translation and dubbing across multilingual contexts \cite{lingua2022}. The system’s pipeline - comprising ASR, MT, and TTS - was optimized via domain-specific training and latency-aware control mechanisms, drawing inspiration from streaming retrieval frameworks \cite{streamrag2023, onlinefusion2021}. Techniques such as pre-aligned script injection and speech pacing control allowed for temporal alignment between transcription and user-uploaded scripts, reducing cascade errors and enabling controllable, user-facing interactivity. These advances resonate with prior efforts in adaptive retrieval for time-constrained generation \cite{realtimeRAG2023}, underscoring a growing emphasis on integrating responsiveness and domain-sensitivity into retrieval-augmented systems.

Concurrently, visual-language research adopted interactive paradigms to enhance interpretability and robustness in classification tasks. Classification by Description pioneered a hybrid framework wherein LLMs generated semantic descriptors - such as “striped fur” or “sharp claws” - to mediate zero-shot image classification \cite{cbd2022}. This approach, akin to attribute-based recognition in early vision work, enabled end-users to comprehend and challenge model decisions via natural language explanations, echoing objectives of explanation-augmented retrieval models like ERA \cite{era2023} and Self-Explainable Transformers \cite{explainabletransformer2022}. The model’s bidirectional interface - both interpreting user intent and exposing model reasoning - highlighted critical challenges in descriptor generalization and bias mitigation, aligning with concerns raised in interpretable RAG systems \cite{intretrieval2023}. These systems collectively mark a trajectory toward retrieval-enhanced architectures that are not only contextually grounded but also semantically transparent and user-adaptive.

By 2023, interactive RAG systems underwent a significant paradigm shift, transitioning from retrieval-dominant pipelines to architectures capable of generating, filtering, and contextualizing their own knowledge scaffolds. GENREAD introduced a generate-then-read framework where LLMs synthesized contextual documents before answer generation, effectively reversing traditional retrieval-first flows \cite{genread2023}. This inversion offered sharper contextual relevance but posed challenges for factual verification, prompting hybrid methods that fused generative document synthesis with post-hoc retrieval-based grounding \cite{realm2020, faithfulnessRAG2023}. Subsequent studies proposed alignment mechanisms to maintain semantic fidelity across generated contexts, balancing autonomy with factual robustness \cite{selfrag2023}.

Concurrently, RAG evolved as a stylistic control mechanism, enabling dynamic modulation of generation tone, complexity, or framing. Retrieval-augmented style transfer models harnessed retrieved exemplars to guide question generation and diversify outputs across pedagogical or creative axes \cite{styletransfer2023, stylisticprompting2022}. Such models reframed RAG not only as an information tool but also as a personalization layer, adapting generation patterns to reflect user preferences, tasks, or domain tones \cite{personaRAG2023}. These advances paralleled explorations in controllable generation via attribute-conditioned decoding and style-aligned fine-tuning \cite{stylecontrol2022}.

Interactive recommendation systems further expanded RAG’s reach into dialog-driven personalization. Chat-REC employed session-aware retrieval to surface contextually aligned recommendations, integrating long-range interaction memory with fine-grained intent modeling \cite{chatrec2023, conversationalrec2022}. Instruction-tuned models like WizardLM complemented this by enhancing multi-turn coherence and rationale generation across diverse information-seeking dialogs \cite{wizardlm2023}. These systems emphasized that retrieval precision must be matched by generative reasoning capabilities to enable transparent and user-controllable outputs \cite{retrievalQA2022, openinstruction2023}.

Simultaneously, structured data interaction became a focal point, with RAG architectures extended to handle tabular, form-based, and programmatic inputs. TableGPT unified table interaction, command execution, and natural language understanding via a shared embedding space \cite{tablegpt2023}. In parallel, form-oriented RAG pipelines facilitated intelligent document automation through semi-structured knowledge alignment and retrieval-enhanced parsing \cite{formrag2023, docprompting2022}. These efforts reflected a shift toward heterogeneous input handling where RAG operates over mixed-modal knowledge structures beyond plain text.

Further sophistication emerged with self-regulatory RAG architectures that equipped LLMs to introspect and revise their retrieval strategies. Self-Debug taught models to evaluate their own retrieval outcomes and issue corrective prompts mid-generation, improving factual alignment and stability \cite{selfdebug2023}. Active RAG conceptualized retrieval not as a static step but a learned policy space - adapting query formulation based on evolving generative states \cite{active-rag2023}. LLM-Embedder proposed a unified representation model for retrieval across memory banks, demonstrations, and external APIs, facilitating agile integration of user-defined contexts \cite{llmembedder2023, retro2021}.

Multimodal and embodied RAG systems closed the loop between perception and response, grounding generation in sensory-rich contexts. MiniGPT-4 exemplified vision-language alignment by enabling RAG pipelines to reference image regions and visual attributes during generation \cite{minigpt42023}. TidyBot extended this further, integrating environmental retrieval cues with robotic actuation for task-aware physical interaction \cite{tidybot2023}. Together, these works redefined RAG not as a preprocessing tool but as a core reasoning substrate, enabling responsive, grounded, and embodied interaction across domains \cite{multirag2023, vidrag2022}.

Interactive RAG systems were also evaluated in high-precision tasks such as math question answering \cite{mathrag2023}, intelligence reporting \cite{fabula2023}, and cross-lingual information access \cite{crosslingualrag2023}. In each case, retrieval was adapted to the user task - optimized for factuality, contextual alignment, or language sensitivity. This adaptability marked a key shift from retrieval-as-lookup to retrieval-as-dialogue, emphasizing the importance of interaction loops that reflect evolving user intent.

The trajectory of RAG in 2024 emphasized deeper alignment between retriever strategies and user intent, broadening its reach into novel domains while refining personalization and adaptive feedback. A persistent challenge was the “preference gap” between retriever scoring metrics and the actual usefulness of retrieved content for LLM-based reasoning. This was directly addressed in Bridging the Preference Gap between Retrievers and LLMs, which proposed relevance estimation grounded in model utility rather than surface similarity \cite{retriever-alignment2024}. Complementary methods such as feedback-based retriever tuning and user-aware scoring functions further narrowed this alignment gap \cite{retrievalalignment2023, reranktuning2024}, enabling retrieval to serve as an active reasoning support layer.

RAG’s role expanded into training-time augmentation, particularly in low-resource settings. RADA introduced retrieval-enhanced data augmentation, constructing synthetic training instances that improved generalization in sparse domains \cite{rada2024}. This built upon prior work in few-shot knowledge injection \cite{kilt2020, mixreview2023}, repositioning RAG not only as an inference-time augmenter but as a curriculum design tool. These approaches demonstrated that LLM performance could be enhanced by shaping its training context with retrieval-informed sampling strategies, especially in biomedical and scientific domains where labeled data is expensive to curate \cite{pubmedrag2023}.

Multimodal RAG pipelines gained traction, integrating heterogeneous signals across vision, text, and audio. mPLUG-Owl showcased a modular architecture enabling cross-modal retrieval fusion, extending applicability to perception-grounded tasks \cite{mplugowl2024}. Educational applications exemplified personalized content delivery: systems tailored programming feedback via lecture materials and supported learning for children with developmental delays using retrieval-adaptive interventions \cite{rag-feedback2024, rag-education2024}. These works paralleled developments in visual-text QA and assistive LLMs, reinforcing RAG’s role in learner-aligned generation \cite{vlm-edtech2023}.

To manage this increasing complexity, modular RAG agents emerged. ERAGent introduced discrete modules for query rewriting, retrieval control, knowledge curation, and reader customization \cite{eragent2024}, echoing trends in agentic LLM systems \cite{rewoo2023}. RA-DIT further optimized this architecture through dual instruction tuning, aligning both retriever and generator to shared task objectives \cite{radit2024}. These frameworks emphasized system-wide alignment, ensuring retrieval and generation operate in a cohesive, goal-aware loop \cite{dualtune2023}.

Multi-turn and persona-aware RAG saw notable advances. IM-RAG modeled internal deliberation chains via inner-monologue-style reasoning between retrieval rounds, improving context coherence in dialogue \cite{imrag2024}. In parallel, systems like PersonaRAG and Adaptive RAG dynamically adjusted retrieval strategies based on user history and profile features, reducing redundancy and enhancing fluency in long-form conversation \cite{personarag2024, adaptiveconvo2024}. These advances complemented exploration in query rewriting and memory-grounded retrieval in dialogue systems \cite{convofusion2023}.

New use cases pushed RAG into content recommendation and multi-perspective search. RAMO and MOOC-RAG enabled adaptive course recommendation through live catalog retrieval and behavior-informed prompts \cite{ramo2024, moocrag2023}. In media domains, RAG enhanced video navigation and summarization across large libraries \cite{ragvideo2024}, while multi-perspective synthesis frameworks used retrieval to surface diverse viewpoints and counterarguments in search and summarization \cite{multiperspective2024, debatellm2023}.

Transparency and evaluation frameworks matured alongside system capabilities. RAG-Ex provided interpretable generation traces, enabling users to understand which retrieved segments influenced output \cite{ragex2024}. In tandem, PromptBench offered a standard benchmarking suite for evaluating retrieval-augmented prompting strategies across datasets and LLMs \cite{promptbench2024}, facilitating comparative assessment and robust reporting. These tools supported explainability and model auditing in production-grade deployments \cite{knowledgetrace2023}.

Over four years, RAG has progressed from static prompting enhancements to sophisticated frameworks enabling dynamic, user-centered interaction. Initial breakthroughs in question generation and retrieval-aware prompting laid the foundation for multimodal, personalized, and feedback-driven systems. Real-world deployments now integrate speech, structured data, and user history, addressing challenges like modality fusion, ambiguity resolution, and proactive alignment. The incorporation of reasoning chains, instruction-aware retrievers, and interpretable outputs positions RAG not merely as a component but as a foundational substrate for human-aligned intelligence \cite{agentrag2024}. This evolution signifies a shift toward interactive systems where retrieval is not a tool but an adaptive cognitive process.

\subsection{RAG for Complex Reasoning and Multi-Step Tasks}
\label{sec:rag-complex-reasoning}

While many early RAG systems focused on factual grounding and information retrieval, complex real-world tasks often demand structured reasoning, multi-step planning, and sequential decision-making. This taxonomy encompasses RAG architectures that explicitly support or enhance such capabilities, either through integration with symbolic structures, iterative generation pipelines, or domain-specific adaptations for reasoning-intensive settings.

A reasoning-aligned objective for Complex Reasoning and Multi-Step RAG systems is formally defined as follows:

\begin{align}
\mathcal{L}_{\text{Reason}} 
&= \sum_{t=1}^{T} \Bigg[
    \underbrace{
    \mathbb{E}_{r_t \sim \mathcal{R}(h_t)}
    \left[ 
        \mathcal{D}_{\text{KL}} \left(
            p_t(y_t \mid h_t, r_t)
            \parallel \hat{y}_t
        \right)
    \right]
    }_{\text{Retrieval-grounded step loss}}
\notag\\
&\quad
    + \lambda \cdot 
    \underbrace{
        \mathcal{C}(r_t, h_t)
    }_{\text{Contextual Coherence}}
    + \gamma \cdot
    \underbrace{
        \mathcal{J}(h_t, h_{t-1})
    }_{\text{Reasoning Consistency}}
\Bigg]
\end{align}

\noindent where:
\begin{itemize}
  \item $T$: Total number of reasoning steps.
  \item $h_t$: Hidden state (or intermediate reasoning output) at step $t$.
  \item $r_t \sim \mathcal{R}(h_t)$: Retrieved documents conditioned on $h_t$ using the retrieval function $\mathcal{R}$.
  \item $p_t(y_t \mid h_t, r_t)$: Model's output distribution at step $t$, conditioned on prior reasoning and retrieved context.
  \item $\hat{y}_t$: Ground-truth or target output at step $t$.
  \item $\mathcal{D}_{\text{KL}}$: Kullback–Leibler divergence measuring discrepancy between the generated and target distribution.
  \item $\mathcal{C}(r_t, h_t)$: Differentiable coherence function quantifying alignment between retrieved context $r_t$ and reasoning state $h_t$.
  \item $\mathcal{J}(h_t, h_{t-1})$: Regularizer penalizing incoherent jumps between successive reasoning states.
  \item $\lambda, \gamma$: Weighting hyperparameters for coherence and reasoning consistency.
\end{itemize}

\noindent This objective explicitly models the multi-hop, interleaved nature of complex reasoning in RAG systems. The first term ensures retrieval-grounded supervision at each reasoning step, the second promotes semantic coherence between retrieved content and reasoning context, and the third enforces logical continuity along the reasoning chain.

\vspace{1em}
\noindent The complete multi-step reasoning pipeline is formalized in Algorithm~\ref{alg:multi-step-rag}.

\begin{algorithm}[H]
\caption{Multi-Step Retrieval-Augmented Reasoning}
\label{alg:multi-step-rag}
\begin{algorithmic}[1]
\Require Initial query $q$, retriever $\mathcal{R}$, generator $\mathcal{G}$, reasoning steps $T$
\Ensure Final output $\hat{y}$
\State $h_0 \gets \text{Encode}(q)$ \Comment{Initial reasoning state from input}
\For{$t = 1$ to $T$}
    \State $r_t \sim \mathcal{R}(h_{t-1})$ \Comment{Retrieve context based on prior state}
    \State $y_t \gets \mathcal{G}(h_{t-1}, r_t)$ \Comment{Generate intermediate output}
    \State $h_t \gets \text{UpdateState}(h_{t-1}, y_t)$ \Comment{Advance reasoning state}
    \State Compute step loss: 
        $\mathcal{L}_t \gets \mathcal{D}_{\text{KL}}(p_t(y_t \mid h_{t-1}, r_t) \parallel \hat{y}_t)$
    \State Add coherence penalty: 
        $\mathcal{L}_t \gets \mathcal{L}_t + \lambda \cdot \mathcal{C}(r_t, h_t)$
    \State Add consistency penalty: 
        $\mathcal{L}_t \gets \mathcal{L}_t + \gamma \cdot \mathcal{J}(h_t, h_{t-1})$
\EndFor
\State \Return Final output $\hat{y} \gets y_T$
\end{algorithmic}
\end{algorithm}

Algorithm~\ref{alg:multi-step-rag} describes a structured reasoning loop in RAG systems designed for multi-hop tasks such as complex question answering, procedural generation, or scientific explanation. The process begins by encoding the query into an initial hidden state $h_0$. At each timestep $t$, relevant documents are retrieved based on the current reasoning state, which then conditions the generator to produce an intermediate output $y_t$. This output updates the hidden state $h_t$, enabling chained inference. Each step is supervised via KL divergence against step-specific ground truths, while coherence and consistency penalties ensure that retrieved content aligns semantically with current reasoning and that the reasoning path remains logically valid across timesteps. This formulation allows RAG systems to move beyond one-shot generation and towards structured, multi-turn, and interpretable reasoning.

Initial innovations in this space emerged from code-related applications, where the structured and logical nature of programming tasks required RAG systems to reason over graphs, syntax trees, and execution patterns. Early studies explored prompt retrieval for knowledge-intensive NLP tasks via models like UPRISE, which dynamically selected prompts to reduce reliance on manual instruction design \cite{uprise2021}. This retrieval-as-guidance framework laid the groundwork for more domain-targeted systems. For instance, {RAG for Code Summarization via Hybrid GNN} introduced a hybrid Graph Neural Network to model local and global code dependencies, integrating retrieval to improve summarization fidelity \cite{codegnn2021}. These works demonstrated that reasoning in structured domains required retrieval pipelines that were not only contextually relevant but also structurally aligned with task-specific logic.

\begin{figure*}[t] 
    \centering
    \includegraphics[width=\textwidth]{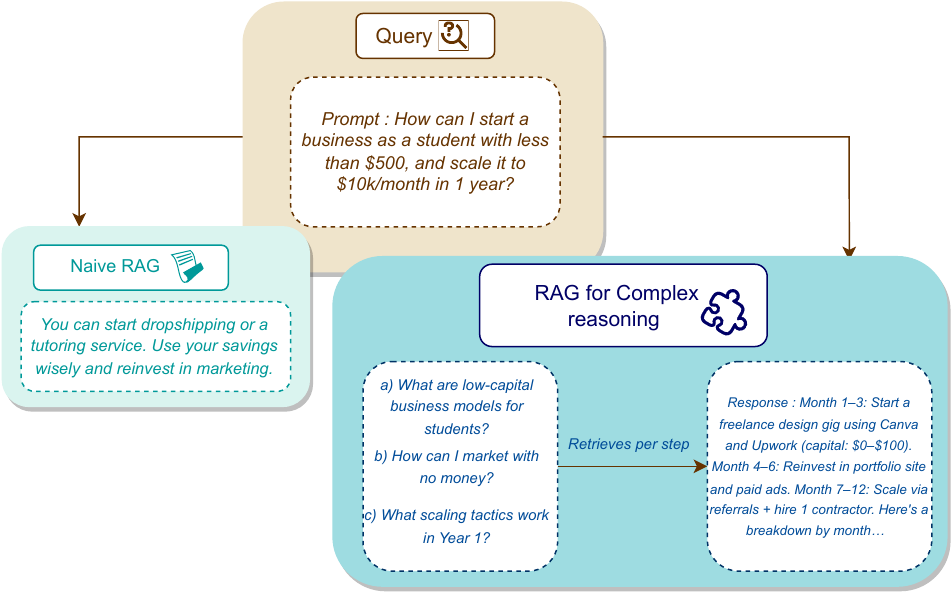} 
    \caption{Illustration of Naïve RAG versus Complex Reasoning RAG. While the Naïve RAG pipeline retrieves documents and generates a direct response to the initial query, the Complex Reasoning RAG decomposes the query into sub-questions, performs step-wise retrieval, and incrementally builds a structured, multi-step response tailored to complex information needs.}
    \label{fig:complex} 
\end{figure*}

Initial innovations in this domain emerged from structured tasks like code generation, where reasoning over syntax trees, dependency graphs, and execution semantics posed challenges for traditional LLMs. UPRISE introduced retrieval-guided prompt selection to reduce dependence on static instruction templates, paving the way for context-sensitive, multi-turn generation \cite{uprise2021}. Building on this, RAG for Code Summarization via Hybrid GNN leveraged a dual-path architecture combining graph neural networks with retrieved code contexts to enhance summarization accuracy and structure awareness \cite{codegnn2021}. REDCODER further advanced this paradigm by framing retrieval as task emulation - modeling how developers retrieve relevant code snippets to inform ongoing edits - thus aligning retrieval with real-world iterative reasoning workflows \cite{redcoder2021}.

RAG methods were soon extended to structured reasoning tasks beyond code, notably in table-based question answering. T-RAG unified dense retrieval and sequence-to-sequence generation into an end-to-end architecture, enabling joint optimization for tabular inputs and reducing reliance on cascaded modules \cite{trag2022}. This architecture improved both factuality and interpretability in open-domain table QA, particularly in settings requiring multi-row inference. Similar efforts in table-text hybrid models further demonstrated the value of tight coupling between structured context retrieval and autoregressive decoding in precision-critical domains \cite{tabllm2022, tablebert2020}.

As applications expanded, inherent reasoning limitations of LLMs became more apparent - especially in planning, counterfactual reasoning, and multi-step inference. Benchmark analyses such as LLMs Still Can’t Plan highlighted common failure modes where models struggled to maintain action coherence across temporal or logical steps \cite{llmplanning2022}. To address this, Automatic Chain-of-Thought prompting introduced retrieval-enhanced step decomposition, guiding models through intermediate reasoning checkpoints prior to final output \cite{cot2022}. This improved performance in weakly supervised environments, especially where symbolic structure or latent dependencies played a role \cite{cotretrieval2023}.

Self-reflective reasoning emerged as a complementary approach. LLMs Can Self-Improve demonstrated iterative refinement mechanisms where models re-evaluated and revised their outputs through self-dialogue loops, effectively mimicking cognitive planning strategies \cite{selfimprove2022}. When paired with external retrieval feedback or persistent memory stores, these systems showed promise for autonomous reasoning progression across task iterations \cite{reflexion2023, scratchpad2022}. Such developments hint at a new class of RAG architectures capable of evolving their own problem-solving pathways through cyclical, feedback-informed reasoning.

Domain-specific adaptations revealed RAG’s versatility in reasoning-intensive contexts. In medicine, retrieval-augmented LLMs demonstrated competence in extracting clinically relevant information with minimal supervision, as shown in retrieval-guided clinical QA frameworks \cite{clinicalrag2022}. These systems leveraged curated medical corpora to supplement incomplete prompts, improving inference reliability and knowledge grounding. Findings from multi-domain retrieval adaptation studies further confirmed that effective RAG requires domain-tailored data representations and query strategies \cite{medqa2021, domadapt2023}.

Conversational agents also benefited from retrieval-guided reasoning. Retrieval-Augmented Response Generation integrated external documents into generation workflows, producing more informative, grounded responses during open-domain dialogue \cite{ragdialogue2022}. By dynamically adjusting retrieval strategies based on user inputs, the system improved contextual alignment and content richness, especially in unpredictable, multi-turn settings \cite{wizardofinternet2023, faithdial2022}. These advances positioned retrieval as a real-time cognitive aid rather than a static fetch mechanism.

By 2023, the field shifted focus to enhancing domain generalization and improving iterative reasoning. Improving Domain Adaptation of RAG Models introduced a multi-domain fine-tuning approach for robust QA across heterogeneous corpora, addressing generalization failures in traditional setups \cite{ragdomain2023}. Complementary research explored dense retriever tuning and unsupervised query reformulation to bolster retrieval utility in zero-shot and few-shot scenarios \cite{gtr2022, universalrag2023}. These works reinforced the need for adaptable pipelines to support flexible deployment across diverse information environments.

Structured reasoning in high-stakes fields remained central. MATHPROMPTER used retrieval to surface relevant formulas and theorems, enabling LLMs to perform multi-step mathematical reasoning over natural language prompts \cite{mathprompter2023}. Medical QA systems like Almanac and ExpertQA integrated retrieval over verified sources to ensure factual accuracy and minimize hallucinations during inference \cite{almanac2023, expertqa2022}. Knowledge Graph-Augmented Language Models further strengthened grounding by replacing free-text retrieval with structured, queryable nodes \cite{kgdialogue2023}.

Several systems moved beyond static retrieval toward adaptive, interleaved strategies. Interleaving Retrieval with Chain-of-Thought enabled context updates at each reasoning step, dynamically supporting evolving thought processes in multi-hop QA \cite{interleavedrag2023}. Retrieve-and-Sample extended this idea by optimizing retrieval with sampling over candidate documents, targeting niche domains like event argument extraction \cite{longtail2023}. Together, these methods reimagined retrieval not as a front-loaded phase but as a looped, on-demand reasoning component.

Emergent reasoning capabilities were increasingly investigated. Analogical reasoning benchmarks showed that when paired with structured retrieval, LLMs begin to mimic human analogy formation, particularly in science QA and concept mapping tasks \cite{analogyrag2023}. LLM+P proposed retrieval-enhanced planning models that aligned external documents with policy-based reasoning, improving task decomposition and sequence fidelity \cite{llmp2023}. Transferable reading comprehension approaches also leveraged RAG to generalize across educational and domain-specific inference settings \cite{adaptreading2023}.

Later efforts refined the retrieval-to-generation pipeline by improving turn-level synchronization and disambiguation. Tree of Clarifications decomposed complex queries into sequential clarifying subquestions, ensuring accurate intent resolution before final generation \cite{clarificationtree2023}. This approach reduced retrieval ambiguity and improved downstream coherence in question answering and task-oriented dialogue \cite{resolvemix2023, iterativeqa2022}. These frameworks underscored the need for tight, interpretable coordination between retrieval dynamics and model reasoning stages.

Novel retrieval paradigms expanded RAG’s adaptability to low-supervision and noisy-data scenarios. Zero-shot retrieval approaches enabled models to identify relevant passages without explicit task-specific retriever training, broadening applicability in open-ended settings \cite{zeroshotrag2023}. Entailment-aware filtering further refined the retrieval process by discarding contradictory or irrelevant context, improving generation faithfulness and consistency \cite{entailmentfiltering2023}. In parallel, PaperQA demonstrated RAG’s value in synthesizing scientific literature, enabling long-context evidence integration for citation-based question answering and literature review workflows \cite{paperqa2023}.

Memory-augmented RAG architectures marked a shift toward longitudinal and reflective reasoning. Self-memory modules allowed LLMs to store and revise past outputs, reinforcing consistency over extended discourse and reducing contradiction across generations \cite{selfmemory2023}. These iterative memory loops resembled deliberative planning, especially when paired with feedback-based refinement mechanisms that guided the model to self-correct based on historical outputs \cite{reflexion2023, memprompt2024}. Together, these strategies hinted at RAG systems capable of evolving answers over time, rather than committing to single-pass inference.

Long-context processing and domain-specific deployment became central to RAG development in 2024. Retrieval Meets Long Context LLMs introduced architectures that combined chunked retrieval with extended context windows, enabling models to process multi-thousand-token documents without degrading performance \cite{longcontextrag2024}. This approach proved critical in domains like academic literature and legal review, where comprehensive grounding is essential. In parallel, medical applications such as preoperative instruction generation illustrated the real-world value of retrieval-enhanced pipelines in supporting patient safety and personalized communication \cite{medinstructionrag2024}. These efforts reinforced the utility of hybrid RAG architectures for grounding LLMs in domain-specific corpora with high precision \cite{medicalrag2023}.

Structured reasoning advanced through graph-guided and precedent-based retrieval systems. G-Retriever and GRAG aligned retrieval paths with underlying graph structures, enabling more accurate response generation in symbolic and relational tasks \cite{gretriever2024, grag2024}. In the legal domain, CBR-RAG combined case-based reasoning with document retrieval, surfacing prior legal decisions based on similarity in query structure and legal framing \cite{CBR-RAG}. These systems mirrored human expert workflows, demonstrating that structured retrieval can support high-accuracy, explainable outputs in rule-governed domains.

Instruction tuning gained traction as a retrieval alignment strategy for complex reasoning tasks. PlanRAG introduced a planning-first pipeline where high-level strategy formulation preceded retrieval, producing more consistent and intention-aligned generations \cite{planrag2024}. InstructRetro fine-tuned retrieval-augmented models with task-specific prompts, optimizing performance for domain-adapted objectives \cite{instructretro2024}. Together with retrieval-aware instruction training frameworks, these methods improved robustness in structured environments requiring long-horizon or multi-turn coherence \cite{retrievalinstruct2023}.

Multi-hop reasoning saw a new wave of innovation driven by feedback-aware and perspective-integrated models. GenGround reversed the traditional retrieval pipeline by generating intermediate reasoning hypotheses first, then validating them through targeted retrieval, creating a flexible loop for complex inference tasks \cite{genground2024}. Unlocking Multi-View Insights expanded this paradigm by integrating diverse perspectives during multi-hop retrieval, enhancing contextual coverage and inference coherence \cite{multiviewrag2024}. These architectures shifted RAG from linear document fetch to iterative, hypothesis-driven exploration \cite{cotretrieval2023}.

Evaluation frameworks evolved to better capture reasoning fidelity and contextual relevance. FACT, FETCH, AND REASON introduced FRAMES, a comprehensive benchmark for assessing accuracy across fact-checking, retrieval quality, and multi-hop inference \cite{frames2024}. Domain-specific evaluations like DomainRAG and Benchmarking RAG for Medicine underscored the need for discipline-adapted metrics and corpora in real-world applications \cite{domainrag2024, medicalbench2024}. Tools like Automated Exam Generation for RAG extended this focus into educational QA, measuring a model’s instructional adaptability and factual grounding under task constraints \cite{exameval2024}.

\begin{table*}[t]
\centering
\caption{Performance metrics of QA-centric Retrieval-Augmented Generation models across multiple evaluation criteria. ``--'' indicates data not reported.}
\label{tab:qa_rag_models}
\begin{tabular}{p{3.5cm}p{3.5cm}p{2.4cm}p{2.4cm}p{2.8cm}}
\toprule
\textbf{Method} & \textbf{Dataset / Setting} & \textbf{Accuracy / F1} & \textbf{Hallucination / Consistency} & \textbf{Retrieval Effectiveness} \\
\midrule
\textbf{FoRAG} & WebGPT (en), WebCPM (zh) & Factuality: 0.82–0.99 & Coherence: 0.91–0.98 & Training Time (Holistic): 33.1h \\
\textbf{WeKnow-RAG} & 4 domains, classification, chunk-size eval & Accuracy: 0.10–0.41 & Hallucination: 0.025–0.35 & Confidence-aware retrieval analysis \\
\textbf{GenGround} & HotpotQA, MuSiQue, StrategyQA & F1: 27.3–52.3 & Semantic Acc: 24.7–55.7 & -- \\
\textbf{FRAMES} & Internal multi-hop QA benchmark & Acc: 0.408–0.729 & -- & Prompting strategy comparison \\
\textbf{Adobe RAG} & Adobe product corpus & nDCG: 0.692–0.822 & -- & Dataset coverage breakdown \\
\textbf{Self-RAG (TA-ARE)} & RetrievalQA (various LLMs) & Match Acc: 6.0–46.4 & -- & Retrieval Acc: up to 100\% \\
\textbf{QA-RAG} & RAPTOR, HFusion, QRFusion & BLEU-1: up to 1.33; METEOR: 0.99 & ROUGE-L: 0.9 (best) & Multi-type QA retrieval fusion \\
\textbf{RQ-RAG} & ARC, POPQA, 2Wiki, MUSIQUE & Acc: 41.7–79.4 & -- & Retrieval Source Comparison (Wiki, DDG, Bing) \\
\textbf{PRCA} & SQuAD, HotpotQA, TopiQCQA & -- & -- & Contextual Adapter across 3 QA datasets \\
\textbf{Face4RAG} & Synthetic + Real QA data (Chinese) & -- & Pos Rate: 30–63\% & Segment-level evaluation \\
\bottomrule
\end{tabular}
\end{table*}

Evaluation frameworks grew increasingly sophisticated, targeting both factual consistency and reasoning fidelity. FACT, FETCH, AND REASON introduced FRAMES, a benchmark that evaluates retrieval-augmented systems across fact-checking, retrieval quality, and multi-hop inference \cite{frames2024}. Complementary efforts such as DomainRAG and Benchmarking RAG for Medicine emphasized the need for field-adapted benchmarks and discipline-specific corpora to assess real-world readiness \cite{domainrag2024, medicalbench2024}. Tools like Automated Exam Generation for RAG explored evaluation in instructional settings, testing a model’s ability to generate grounded, curriculum-aligned responses in education-oriented QA \cite{exameval2024}.

Security and robustness became pressing concerns as RAG pipelines matured. Black-Box Opinion Manipulation Attacks revealed how adversarially selected documents could skew generation outputs, raising concerns about content injection vulnerabilities \cite{blackboxattack2024}. To mitigate such risks, new retrieval mechanisms were introduced - such as missing-information-guided querying and entailment-aware filters - to improve resilience under incomplete or deceptive input settings \cite{missinginfo2024}. DR-RAG further introduced dynamic relevance scoring techniques to ensure stable generation even in the presence of noisy or adversarial retrieval artifacts \cite{drrag2024}.

Novel real-world deployments showcased RAG’s practical versatility across complex, high-stakes domains. SAPPhIRE-RAG brought structured reasoning to engineering design workflows by linking functional models with adaptive retrieval \cite{sapphirerag2024}. RichRAG demonstrated multi-intent generation capabilities, enabling systems to address overlapping or evolving user needs in single queries \cite{richrag2024}. In education, Lecture-RAG facilitated feedback generation from instructional content, while i-MedRAG integrated iterative follow-ups in clinical QA pipelines, refining medical responses through multi-turn clarification loops \cite{lecturefeedbackrag2024, imedrag2024}.

Over four years, RAG has transformed from modular augmenters to cognitively aligned reasoning frameworks. Early foundations in prompt selection and structured retrieval evolved into complex pipelines capable of symbolic reasoning, domain-specific adaptation, and iterative correction. Innovations like graph-aligned search, memory-driven refinement, and multi-perspective synthesis positioned RAG not merely as an augmentation layer, but as a reasoning substrate underpinning trustworthy AI \cite{grag2024, reflexion2023, multiviewrag2024}. This paradigm shift - from retrieval-enhanced response to retrieval-mediated cognition - places RAG at the core of adaptive, explainable, and reliable language-based intelligence.

\section{Applications of RAG-LLMs}
\label{sec:applications}

\subsection{Information Retrieval and Question Answering}
\label{sec:rag-qa}

Open-domain Question Answering (ODQA) has seen significant advancements through the integration of Retrieval-Augmented Generation (RAG) techniques with Large Language Models (LLMs). Recent research has focused on improving the accuracy, factuality, and coherence of generated answers. FoRAG addresses factual inaccuracy and logical coherence in long-form question answering through an outline-enhanced generator and a doubly fine-grained RLHF framework~\cite{forag}. WeKnow-RAG combines Web search and Knowledge Graphs with RAG systems to enhance accuracy and reduce hallucinations, employing multi-stage retrieval and self-assessment mechanisms~\cite{weknowrag}. The Generate-then-Ground (GenGround) framework tackles multi-hop question answering by alternating between answer deduction and knowledge grounding phases, improving upon traditional retrieve-then-read paradigms~\cite{genground}.

Recent advancements in Retrieval-Augmented Generation (RAG) systems have led to the development of comprehensive evaluation frameworks and datasets. FRAMES (Factuality, Retrieval, And reasoning MEasurement Set) offers a unified approach to assess Large Language Models' (LLMs) factual accuracy, retrieval capabilities, and reasoning skills through challenging multi-hop questions~\cite{frames}. A domain-specific framework for Adobe products addresses the inefficiency of general pre-trained models in specialized contexts by leveraging a smart retrieval system trained on product-specific data and user behavior~\cite{domainrag}. RetrievalQA, a benchmark of 1,271 questions, evaluates Adaptive RAG (ARAG) methods and introduces Time-Aware Adaptive Retrieval (TA-ARE) to enhance retrieval decisions without additional training~\cite{retrievalqa}. These contributions collectively aim to improve RAG systems' performance, efficiency, and applicability across various domains. Future research directions include refining the integration of knowledge sources, enhancing self-assessment mechanisms, and expanding the applicability of these systems to more diverse and dynamic information environments, ultimately advancing the field of Open-domain Question Answering.

Recent advancements in Retrieval-Augmented Generation (RAG) systems have focused on improving response quality, factual consistency, and query refinement. QA-RAG introduced a novel approach that reformats retrieved chunks into question-answer pairs, enhancing context representation and reducing hallucinations~\cite{qarag}. RQ-RAG developed a framework for dynamic query refinement, addressing limitations in handling complex queries and improving performance on single-hop and multi-hop question-answering tasks~\cite{rqrag}. PRCA proposed a Pluggable Reward-Driven Contextual Adapter to refine retrieved information before feeding it into black-box large language models, enhancing ReQA performance without fine-tuning the entire model~\cite{prca}. Face4RAG introduced a comprehensive benchmark for evaluating factual consistency in Chinese RAG systems, along with L-Face4RAG, a method for detecting logical fallacies~\cite{face4rag}. These innovations collectively aim to improve the accuracy, relevance, and factual consistency of RAG systems across various domains and languages. Future research directions include optimizing retrieval-generation alignment, exploring broader applications, and refining methods for diverse error distributions and languages.

\begin{table*}[t]
\centering
\renewcommand{\arraystretch}{1.15} 
\caption{Overview of Security, Fairness, and Privacy-focused RAG Systems}
\begin{tabular}{@{}lll@{}}
\toprule
\textbf{Model} & \textbf{Datasets / Evaluation Context} & \textbf{Key Metrics / Findings} \\
\midrule
PoisonedRAG & NQ, HotpotQA, MS-MARCO & ASR: 0.97–0.99; F1-Score: 0.96–1.00 \\
RAG (CoCondenser + MiniLM) & Government, Education, Society, Health & ASR: 0.17–0.50; ASV: –0.17 to 0.67 \\
RC-RAG & Internal eval on ChatGPT and Mistral & Risk: 14.94–19.00; Carefulness: 52.87–65.37 \\
Towards Fair RAG & Exposure Disparity Benchmarks & EE-D: 0.14; EE-R: 0.28 \\
LLaMA2-7B-Chat & Health, Enron Attacks & ROUGE Prompts: 73–111; Repeat Contexts: 55–135 \\
BadRAG & GPT-4, Claude-3 Evaluation & Retrieval Success: 98.9\%; Rejection Rate: 74.6\% \\
SAGE & HealthcareMagic, Wiki-PII & BLEU-1: 0.01–0.11; ROUGE-L: 0.02–0.09 \\
\bottomrule
\end{tabular}
\label{tab:security_fairness_rag}
\end{table*}

Table~\ref{tab:qa_rag_models} presents a consolidated view of RAG models designed for open-domain and multi-hop question answering. Models such as FoRAG and QA-RAG demonstrate strong factuality and BLEU-based performance, respectively, while frameworks like WeKnow-RAG emphasize hallucination mitigation and confidence calibration. GenGround and RQ-RAG highlight the growing importance of multi-hop and interleaved retrieval strategies, with tailored designs for complex question decomposition. Meanwhile, Self-RAG and TA-ARE showcase improvements in retrieval reliability through thresholding and time-aware adaptation. Across benchmarks like HotpotQA, StrategyQA, SQuAD, and domain-specific corpora, the results reflect a clear trend: recent RAG architectures increasingly prioritize factual accuracy, retrieval-query alignment, and context-aware generation. This table underscores the diversity in model design and evaluation strategies across the QA-centric RAG landscape.

Looking ahead, future research in RAG-based QA must address several key challenges: refining the integration of heterogeneous knowledge sources, improving the robustness of query reformulation, and ensuring consistent factual grounding across diverse domains. As RAG systems continue to evolve, their application in real-time, domain-sensitive information retrieval will play a central role in building trustworthy and high-utility QA systems.

\subsection{Security, Fairness, and Privacy in RAG}
\label{sec:rag-security}

As RAG systems become increasingly integrated into real-world applications, concerns surrounding security, fairness, and privacy have emerged as critical areas of investigation. RAG’s reliance on external retrieval introduces new attack surfaces and ethical challenges, particularly in high-stakes domains such as healthcare, law, and public information systems.

Recent work has exposed vulnerabilities to knowledge corruption and adversarial manipulation. Studies such as {PoisonedRAG}~\cite{poisonedrag} and {BadRAG}~\cite{badrag} demonstrate that malicious actors can influence generated outputs through strategically inserted documents in the retrieval corpus - achieving high manipulation success rates with minimal data injection. Black-box opinion manipulation attacks~\cite{blackboxopinion} further reveal how content polarity and factual framing can be covertly altered via adversarial retrieval inputs.

To defend against such threats, counterfactual prompting frameworks~\cite{counterfactualprompting} have been proposed, which assess model uncertainty by simulating degraded retrieval contexts and filtering low-confidence generations. Complementary work has focused on fine-grained access control for retrieval systems~\cite{ragaccesscontrol}, ensuring that sensitive information remains accessible only to authorized users through on-premise deployments and document-level filtering.

Privacy studies~\cite{privacyissuesrag} have shown a dual-sided role of RAG: while retrieval systems may leak private data under certain configurations, they can also mitigate over-reliance on memorized training data in LLMs. Frameworks like {SAGE} (Synthetic Attribute-based Generation with Agent-based refinement)~\cite{sageframework} offer synthetic data generation pipelines that balance contextual richness with privacy preservation - critical for healthcare and other regulated industries.

Fairness remains a pressing concern, particularly in ranking algorithms used within RAG pipelines. Biases in exposure, document selection, and model responses can reinforce systemic inequities. Integrating fair ranking algorithms during retrieval~\cite{fairrag} and developing multilingual, bias-aware QA systems are active research directions.

The empirical results summarized in Table~\ref{tab:security_fairness_rag} highlight the multifaceted vulnerabilities and control strategies across RAG systems. PoisonedRAG and BadRAG illustrate high success rates for knowledge corruption attacks, especially in open-domain QA tasks. Meanwhile, fairness-focused frameworks like {Towards Fair RAG}~\cite{fairrag} emphasize exposure disparity metrics, showcasing efforts to rebalance content exposure. Privacy-oriented systems like {SAGE}~\cite{sageframework} demonstrate trade-offs between utility (e.g., BLEU, ROUGE) and information leakage control across datasets. The inclusion of metrics such as ASR, ASV, NDCG variation, and targeted information leaks also reveals a growing trend toward quantifying both adversarial risk and protection efficacy in RAG pipelines~\cite{poisonedrag,badrag,privacyissuesrag,blackboxopinion}.

Overall, the diversity in attack types, evaluation setups, and defense metrics reflects the field's push toward comprehensive security and fairness guarantees in RAG-enabled systems.

Future challenges include improving the resilience of RAG pipelines to adversarial and noisy inputs~\cite{counterfactualprompting}, extending privacy safeguards with formal guarantees (e.g., differential privacy)~\cite{sageframework}, and embedding fairness constraints throughout the retrieval-to-generation workflow~\cite{fairrag}. Building trustworthy RAG systems will require a combination of robust retrieval architectures, interpretable generation mechanisms, and ethical alignment across all components of the pipeline.

\subsection{Conversational Agents and Chatbots}
\label{sec:rag-chatbots}

\begin{table*}[t]
\centering
\small
\renewcommand{\arraystretch}{1.15}  
\caption{Evaluation of Conversational and User-Centric RAG Systems}
\begin{tabular}{l l l l}
\toprule
\textbf{Model} & \textbf{Evaluation Category} & \textbf{Metric} & \textbf{Score} \\
\midrule
\multirow{4}{*}{RAGate} 
    & Retrieval Performance & TF-IDF Recall@1 / @3 & 0.0227 / 0.0871 \\
    & Retrieval Performance & BERT-Ranker Recall@1 / @3 & 0.2475 / 0.4714 \\
    & Adaptive Augmentation (Zero-shot) & Llama-2-13B F1 & 0.1230 \\
    & PEFT - Context Only & Precision / Recall / F1 & 0.5203 / 0.3359 / 0.4082 \\
\hdashline[1pt/1pt]
\multirow{3}{*}{PersonaRAG} 
    & TriviaQA Accuracy & Top-3 / Top-5 & 94.73 / 89.47 \\
    & WebQ Accuracy & Top-3 / Top-5 & 63.46 / 67.50 \\
    & NQ Accuracy & Top-3 / Top-5 & 49.02 / 48.78 \\
\hdashline[1pt/1pt]
\multirow{4}{*}{TidyBot} 
    & Summarization Accuracy & Unseen Objects & 91.2\% \\
    & Execution Success Rate & Primitive Execution & 96.2\% \\
    & Localization Accuracy & Localized Objects & 95.5\% \\
    & Cleanup Speed & Time per Object & 15-20 seconds \\
\hdashline[1pt/1pt]
\multirow{3}{*}{GPT-4-32k} 
    & MRR (Simplified Qns) & Version-specific & 0.295 \\
    & Recall@20 (Original Qns) & Version / Non-version & 0.5 / 0.6 \\
    & Preliminary Eval MRR & Simplified / Original & 0.424 / 0.383 \\
\bottomrule
\end{tabular}
\label{tab:conversational_eval}
\end{table*}

RAG has emerged as a powerful framework for enhancing conversational agents by enabling dynamic access to external knowledge during dialogue generation. Recent developments in this area have focused on improving personalization, contextual adaptation, and retrieval efficiency - addressing the limitations of static knowledge and non-specific responses in traditional LLM-based systems.

A major innovation in adaptive knowledge integration is {RAGate}~\cite{ragate}, which proposes a gating mechanism that determines when and how to augment conversational responses with external retrievals. By analyzing the semantic and pragmatic properties of dialogue turns, RAGate selectively invokes the RAG module only when necessary, avoiding over-reliance on external data that can result in incoherent or irrelevant outputs. Its three variants - Prompt-based (RAGate-Prompt), Parameter-Efficient Fine-Tuning (RAGate-PEFT), and Multi-Head Attention-based (RAGate-MHA) - offer complementary strategies for controlling retrieval usage in dynamic conversations.

Personalization has also become a central focus of RAG-enabled conversational systems. {PersonaRAG}~\cite{personarag} introduces a user-centric agent framework that tailors both retrieval and generation to the user’s evolving preferences and dialogue history. It incorporates modular agents such as User Profile, Contextual Retrieval, and Feedback Agents to enable dynamic adaptation and continuous personalization. This marks a critical step toward more interactive, user-aware conversational AI.

Applications of RAG in real-world embodied agents are exemplified by {TidyBot}~\cite{tidybot}, which uses LLMs to learn user-specific household organization patterns. By summarizing user behavior and inferring generalized object placement preferences, TidyBot~\cite{tidybot} demonstrates how RAG frameworks can support grounded, context-sensitive assistance in physical environments.

In customer support settings, RAG-based chatbots have been optimized through improved keyword extraction, version-aware retrieval, and prompt design~\cite{customersupportrag}. These enhancements lead to more accurate resolution of technical queries and personalized troubleshooting recommendations, particularly in knowledge-dense support workflows.

The experimental results from Table~\ref{tab:conversational_eval} indicate the increasing sophistication of RAG-based conversational agents. RAGate demonstrates significant improvements in adaptive augmentation, particularly when fine-tuned using parameter-efficient techniques and multi-head attention. PersonaRAG ~\cite{personarag} outperforms across datasets like TriviaQA ~\cite{triviaqa} and WebQ ~\cite{webq}, highlighting the effectiveness of personalized agents in retrieval and response synthesis. TidyBot's~\cite{tidybot} performance in real-world scenarios confirms the viability of task-specific LLM-based assistants, achieving over 90\% accuracy in localization and summarization tasks. Finally, evaluations on GPT-4-32K reinforce the importance of version-specific information retrieval, with improved mean reciprocal rank scores and recall across both simplified and original question formulations.

Future directions include advancing multi-turn semantic understanding, improving response consistency under dynamic knowledge updates, and developing stronger alignment between user intent and retrieval outputs. Moreover, privacy-preserving retrieval mechanisms and adaptive personalization algorithms will be key to scaling RAG-based chatbots in real-world, user-sensitive applications.

\subsection{Retrieval-Based Code Generation}
\label{sec:rag-codegen}

The application of RAG to code generation and synthesis has introduced new capabilities in handling domain-specific languages, adapting to rare programming patterns, and improving generation coherence across diverse codebases. These systems augment LLMs with task-relevant code snippets, documentation, and usage examples - enabling more accurate and context-aware code completion, particularly in zero-shot or few-shot settings.

One such system is {ProCC}~\cite{procc}, which combines prompt-based retrieval with a contextual multi-armed bandit algorithm to dynamically select among multiple semantic perspectives of code. By leveraging a multi-retriever setup and adaptive retrieval strategies, ProCC~\cite{procc} enhances the alignment between user intent and retrieved contexts. It achieves significant performance gains over existing code completion baselines, demonstrating strong generalization to both open-source and proprietary codebases.

\begin{table*}[t]
\centering
\small
\renewcommand{\arraystretch}{1.15} 
\caption{Comparison of Retrieval-Based Code Generation Systems and Metrics}
\label{tab:codegen_eval}
\begin{tabular}{l l l l}
\toprule
\textbf{Model} & \textbf{Category} & \textbf{Metric} & \textbf{Score / Description} \\
\midrule
\multirow{10}{*}{GitHub Copilot} 
  & IDE Support & Supported IDEs & IntelliJ, VSCode, PyCharm, etc. \\
  & Reference / Explanation & Provides References & No \\
  & Reference / Explanation & Explanation of Suggestions & No \\
  & Suggestion Variety & Number of Suggestions & Up to 10 \\
  & Training Source & Data & Public Repositories \\
  & Languages Supported & Best With & C, C++, Java, Python, etc. \\
  & Accessibility & Can Be Used Offline & No \\
  & Accessibility & Can Access Local Files & Yes \\
  & Release Info & Developer / Release & OpenAI-Microsoft / Oct 2021 \\  & Pricing & Subscriptions & Free for Students, $10–$19/month \\
\hdashline[1pt/1pt]
\multirow{6}{*}{Amazon CodeWhisperer} 
  & IDE Support & Supported IDEs & JetBrains, VS Code, AWS Cloud9 \\
  & Reference / Explanation & Provides References & Yes \\
  & Suggestion Variety & Number of Suggestions & Up to 5 \\
  & Training Source & Data & Public Code \\
  & Accessibility & Offline / Local Files & No / Yes \\
  & Release Info & Developer / Release & AWS / June 2022 \\
\hdashline[1pt/1pt]
\multirow{6}{*}{ChatGPT} 
  & IDE Support & Supported IDEs & None \\
  & Reference / Explanation & Explanation of Suggestions & Yes \\
  & Suggestion Variety & Number of Suggestions & One (manual repeat) \\
  & Training Source & Data & GitHub, GitLab, Codex \\
  & Accessibility & Offline / Local Files & No / No \\
  & Release Info & Developer / Release & OpenAI / Nov 2022 \\
\hdashline[1pt/1pt]
\multirow{4}{*}{ProCC} 
  & Accuracy & EM (Open-Source / Private) & 8.6\% / 10.1\% \\
  & Fine-Tuning Gain & Fine-Tuned Model Improvement & 5.6\% \\
  & Model Performance & CodeLlama EM / ES & 54.66 / 75.85 \\
  & Model Performance & StarCoder EM / ES & 49.14 / 72.69 \\
\hdashline[1pt/1pt]
\multirow{3}{*}{RAG + Few-Shot for DSLs} 
  & GPT-3.5 Error Rate & 1–4 Shots & 100\%, 100\%, 96\%, 73\% \\
  & GPT-4 Error Rate & 1–4 Shots & 40\%, 49\%, 42\%, 29\% \\
\hdashline[1pt/1pt]
\multirow{8}{*}{CodeT5+} 
  & Code Completion & pass@1 / @10 / @100 & 35.0\%, 54.5\%, 77.9\% \\
  & Math QA (Python) & pass@80 / @100 & 87.4\%, 73.8\% \\
  & GSM8K (Python) & pass@80 / @100 & 73.8\%, 87.4\% \\
  & Summarization & BLEU-4 & 33.83\% \\
  & Completion & Exact Match & 44.86\% \\
  & Retrieval & MRR (Text-to-Code) & 77.4 \\
  & Defect Detection & Accuracy & 66.7\% \\
  & Clone Detection & F1 Score & 95.0\% \\
\bottomrule
\end{tabular}
\end{table*}

Retrieval has also proven valuable in synthesizing programs for rare or domain-specific languages. A combined RAG and Few-Shot Learning (FSL)~\cite{dslragfsl} approach has been used to generate models for uncommon domain-specific languages (DSLs) ~\cite{dsl}, such as the MontiCore Sequence Diagram Language. This method retrieves similar examples from relevant DSL corpora, enabling LLMs to generate correct syntax and semantics with minimal prior exposure - offering a scalable strategy for supporting niche development environments.

{CodeT5+}~\cite{codet5plus}, a flexible encoder-decoder architecture, extends the applicability of code generation models by supporting multiple modalities and training objectives. It incorporates span denoising, contrastive learning, text-code matching, and causal language modeling, resulting in a unified model that performs strongly across a range of understanding and generation tasks. CodeT5+ exemplifies how retrieval-enhanced training regimes can improve downstream generalization and adaptability.

Table IV presents a comparative overview of popular retrieval-based and retrieval-enhanced code generation systems. It reveals how foundational systems like GitHub Copilot and CodeWhisperer prioritize integration and accessibility, while newer models such as ProCC and CodeT5+ show significant gains in exact match accuracy and task-specific performance. Notably, retrieval-enhanced few-shot synthesis for uncommon DSLs exhibits reduced error rates with GPT-4 as shot count increases, indicating its promise in low-resource settings. The diversity of metrics - from BLEU to pass@k, and clone detection F1 - highlights the multidimensional nature of evaluating such systems. These comparisons reinforce the importance of retrieval and task-aware adaptation in enhancing LLM capabilities for program synthesis, summarization, and reasoning across diverse code domains.

Future work in this area includes improving the quality of retrieval in noisy or sparsely annotated repositories, enabling on-the-fly updates to retrieval indices for dynamic codebases, and ensuring privacy in proprietary environments. Additional goals involve bridging the performance gap between open-source and closed-source models, improving reasoning over multi-file and multi-library contexts, and unifying RAG with program verification and synthesis frameworks to ensure both correctness and usability in generated code.

\subsection{Educational and Corporate Use Cases}
\label{sec:rag-education-corporate}

\begin{table*}[t]
\centering
\renewcommand{\arraystretch}{1.15} 
\caption{Evaluation of RAG-based and LLM-enhanced systems in educational and corporate contexts}
\label{tab:edu_corp_usecases}
\resizebox{\linewidth}{!}{%
\begin{tabular}{@{}lp{5cm}p{5cm}@{}}
\toprule
\textbf{Model} & \textbf{Metric} & \textbf{Details} \\
\midrule
\multirow{3}{*}{LLaMA2} 
& Precision & 89\% \\
& Recall & 84.5\% \\
& Accuracy & 85\% \\
\hdashline[1pt/1pt]
\multirow{8}{*}{RAG (Cost across prompting strategies)} 
& Zero-shot Prompt Type I (GPT-3.5 Turbo) & \$0.100 \\
& Zero-shot Prompt Type II (GPT-3.5 Turbo) & \$0.014 \\
& Tree of Thoughts Prompt (GPT-3.5 Turbo) & \$0.013 \\
& RAG Prompt (GPT-3.5 Turbo) & \$0.008 \\
& Zero-shot Prompt Type I (GPT-4 Turbo) & \$1.035 \\
& Zero-shot Prompt Type II (GPT-4 Turbo) & \$0.188 \\
& Tree of Thoughts Prompt (GPT-4 Turbo) & \$0.137 \\
& RAG Prompt (GPT-4 Turbo) & \$0.137 \\
\hdashline[1pt/1pt]
\multirow{7}{*}{RAG with GPT-4 (Lecture Feedback System)} 
& Avg. Feedback Time (with lecture) & 18 seconds \\
& Avg. Feedback Time (without lecture) & 1–2 seconds \\
& Linked Lecture Segments & 160 segments \\
& Avg. Segments per Feedback & 1.67 \\
& Student Satisfaction (Length) & Neutral to Agree \\
& Student Trust (Accuracy) & Neutral to Agree \\
& System Usability Score & 74.8 \\
\hdashline[1pt/1pt]
Xwin-LM-70B-V0.1-GPTQ 
& Question-wise Accuracy & Q1–Q6: 100\%, Overall: 100\% \\
\bottomrule
\end{tabular}%
}
\end{table*}

RAG-LLMs are increasingly being adopted in educational contexts to support personalized tutoring, automated feedback, and intelligent content generation. By combining retrieval capabilities with LLMs’ generative strengths, RAG enables adaptive learning systems that respond to individual needs and contextual learning gaps.

In automated tutoring, systems have employed RAG to assess social-emotional competencies in tutors, generate personalized feedback for programming learners, and assist children with developmental disabilities. For instance, {PicotTo} integrates offline LLMs with RAG-based retrieval to deliver tailored educational content to children with learning challenges~\cite{picotto}. Similarly, frameworks like {RAST} have been proposed for automated question generation, enabling the creation of diverse and targeted exam preparation materials - an essential advancement in scalable educational tooling~\cite{rast}.

Future directions in educational RAG include integrating more sophisticated retrieval pipelines to support multi-turn dialogue-based learning, expanding RAG’s role in formative and summative assessments, and refining alignment between pedagogical goals and retrieved content. As educational institutions adopt hybrid learning environments, RAG systems offer a promising avenue for delivering real-time, context-aware, and equitable educational support.

In corporate settings, RAG is being leveraged for intelligent knowledge management across heterogeneous data sources. Recent work has focused on optimizing retrieval from domain-specific corpora through embedding fine-tuning, vector space segmentation, and LLM-based re-ranking~\cite{rag-dataspaces}. These techniques reduce hallucination in high-stakes business queries and improve contextual accuracy in enterprise search.

Innovations such as element-based chunking for financial documents~\cite{financial-chunking} and automated form-filling pipelines~\cite{form-filling} showcase RAG’s ability to process structured and semi-structured data, supporting workflows in legal, financial, and operational domains. Additionally, multi-vector retrieval strategies have demonstrated effectiveness in consolidating fragmented knowledge across business units.

Table~\ref{tab:edu_corp_usecases} provides a comparative overview of retrieval-augmented systems and LLMs deployed in educational and enterprise settings. The LLaMA2 baseline exhibits strong accuracy for basic classification tasks, while RAG-enhanced GPT-4 models show substantial benefits in feedback systems, particularly when grounded in lecture content. Cost-wise, RAG-based prompting emerges as significantly more economical compared to zero-shot or structured reasoning approaches, especially for large models like GPT-4. The user feedback metrics - such as student satisfaction, trust, and usability - demonstrate growing acceptance of such systems. {Xwin-LM-70B}'s consistent performance across all test questions suggests the feasibility of using high-capacity models for instructional support and knowledge management. These findings highlight the practical value of retrieval grounding and personalization in building scalable, cost-efficient educational and corporate AI systems.

The results summarized in Table~\ref{tab:corporate_rag_metrics} reveal the growing maturity of RAG systems for corporate knowledge management. The {RAST} framework demonstrates measurable BLEU improvements across QA datasets, suggesting the utility of style-aware augmentation. The {bge-large-en-v1.5} model achieves consistently high accuracy across separate and combined vector spaces, underlining the effectiveness of embedding-level architecture in complex document corpora. Meanwhile, {DS-RAG} emphasizes flexibility in handling multi-provider and policy-aware environments, a key requirement for large-scale enterprise deployment.

The most granular breakdown is offered by {Chipper}, which evaluates RAG’s behavior over financial documents~\cite{financial-chunking}. Its multi-resolution chunking analysis illustrates the trade-offs between chunk size and retrieval quality, while its Q\&A experiments demonstrate how chunking strategy impacts both generative accuracy and answer completeness. Notably, summary-based and prefix-informed chunking outperformed standard baselines, particularly when grounded against GPT-4 and human references. These findings reinforce the importance of hybrid chunking strategies and structured content modeling in enterprise-grade RAG deployments.

Ongoing research explores using LLMs as early-stage classifiers for query routing, refining data comparison techniques for more effective chunk selection, and improving retrieval efficiency in time-sensitive decision pipelines. As organizations continue to scale their digital infrastructure, RAG offers a robust framework for transforming enterprise knowledge retrieval into an intelligent, context-aware system.

\subsection{Scientific and Legal Text Processing}
\label{sec:rag-scientific-legal}

\begin{table*}[t]
\centering
\renewcommand{\arraystretch}{1.15} 
\caption{Metrics and Evaluation Results for RAG-based Corporate Knowledge Management Systems}
\label{tab:corporate_rag_metrics}
\begin{tabular}{@{}llp{10cm}@{}}
\toprule
\textbf{Model} & \textbf{Metric} & \textbf{Values and Evaluation Details} \\
\midrule
\multirow{6}{*}{RAST (Style Transfer)} & BLEU Scores (SQuAD/1) & Top-1: 19.25, Oracle: 23.23, Pairwise: 48.91, Overall: 9.14 \\
& BLEU Scores (SQuAD/2) & Top-1: 19.36, Oracle: 22.59, Pairwise: 56.42, Overall: 7.75 \\
& BLEU Scores (NewsQA) & Top-1: 11.02, Oracle: 16.26, Pairwise: 23.16, Overall: 7.74 \\
\hdashline[1pt/1pt]
\multirow{6}{*}{bge-large-en-v1.5} & NDCG@10 & 0.85 \\
& Accuracy@k (Combined vs. Separate DBs) & k=1: 0.85 / 0.91, k=5: 0.92 / 0.95, k=10: 0.93 / 0.97 \\
& Training/Test Split & Written: 1450 / 500, Transcribed: 1450 / 500 \\
& Corpus & Articles: 165 (2004 chunks), Transcribed Data: 59 (4167 chunks) \\
\hdashline[1pt/1pt]
\multirow{4}{*}{DS-RAG} & Retrieval Latency & Network-dependent \\
& Embedding Consistency & Same across providers \\
& Ranking Time & Tech-dependent \\
& Similarity Accuracy & High (model-dependent) \\
\hdashline[1pt/1pt]
\multirow{14}{*}{Chipper (Financial Reports)} & Element Distribution & NarrativeText: 61,780; Title: 29,664; Tables: 7,700; etc. \\
& Chunking Stats & Base128: 64k chunks (Mean 800), Base256: 32k, Base512: 16k, Chipper: 20.8k \\
& Retrieval Accuracy (Page Level) & Base128: 72.34\%, Base256: 73.05\%, Base512: 68.09\%, Aggregated: 83.69\% \\
& Retrieval ROUGE/BLEU & Base128: 0.383 / 0.181, Base256: 0.433 / 0.231, Base512: 0.455 / 0.250 \\
& Chipper Aggregated Scores & ROUGE: 0.568, BLEU: 0.452, Accuracy: 84.40\% \\
& Keywords Chipper Scores & ROUGE: 0.444, BLEU: 0.315, Accuracy: 46.10\% \\
& Summary Chipper Scores & ROUGE: 0.473, BLEU: 0.350, Accuracy: 62.41\% \\
& Prefix+Table Chipper Scores & ROUGE: 0.514, BLEU: 0.400, Accuracy: 67.38\% \\
& Q\&A Results (Base128) & No answer: 35.46\%, GPT-4: 29.08\%, Manual: 35.46\% \\
& Q\&A Results (Base256) & No answer: 25.53\%, GPT-4: 32.62\%, Manual: 36.88\% \\
& Q\&A Results (Base512) & No answer: 24.82\%, GPT-4: 41.84\%, Manual: 48.23\% \\
& Q\&A Results (Keywords) & No answer: 22.70\%, GPT-4: 43.97\%, Manual: 53.19\% \\
& Q\&A Results (Summary) & No answer: 17.73\%, GPT-4: 43.97\%, Manual: 51.77\% \\
& Q\&A Results (Prefix+Table) & No answer: 20.57\%, GPT-4: 41.13\%, Manual: 53.19\% \\
\bottomrule
\end{tabular}
\end{table*}

Scientific and legal domains demand high-precision, contextually grounded language processing---a challenge that RAG is well-positioned to address. Recent work in RAG and Retrieval-Augmented Understanding (RAU) has enabled models to reason over dense, domain-specific texts while reducing hallucination and improving explainability.

In legal applications, {CBR-RAG} combines case-based reasoning with RAG, allowing LLMs to anchor answers in legally relevant precedents through domain-specific embeddings~\cite{cbr}. {LegalBench-RAG} introduces a benchmark of 6,858 query-answer pairs, designed to test retrieval fidelity at a fine-grained level across a massive corpus of legal documents~\cite{legalbench-rag}. To support policy-related inference, {HyPA-RAG} employs hybrid dense-sparse-KG retrieval and adaptive query complexity tuning, improving the alignment of generated outputs with legal and regulatory standards~\cite{hypa-rag}.

Medical and scientific domains have seen parallel innovations. {MVRAG} proposes multi-perspective query rewriting to improve retrieval accuracy in healthcare and research domains~\cite{mvrag}, while RAG systems for clinical QA (e.g., for generating preoperative instructions) have achieved accuracy rates exceeding 90\%. These systems provide context-aware guideline generation and efficient patient communication tools. In scientific education, {RAMO} applies RAG to cold-start recommendation problems in MOOCs, offering adaptive course suggestions based on minimal user history~\cite{ramo}.

Further research has emphasized retrieval alignment in long-context documents, adversarial robustness, and the integration of domain-specific evaluation metrics. The survey on Retrieval-Augmented Language Models ({RALMs})~\cite{ralm-survey} provides a unified overview of how RAG techniques are being embedded into both generative and comprehension pipelines across professional domains.

Table~\ref{tab:legal_sci_rag_1} and Table~\ref{tab:legal_sci_rag_2} summarize the performance of retrieval-augmented models across diverse scientific and legal tasks. {Forms Copilot} shows how different prompting strategies affect factuality and coherence in structured document generation. {MVRAG}~\cite{mvrag} and {CaseGPT}~\cite{casegpt} highlight the growing emphasis on precision in legal and medical domains, with {CaseGPT} achieving high NDCG and MRR in both settings~\cite{casegpt}. Meanwhile, {GPT-4.0-RAG} outperforms fine-tuned baselines in hallucination reduction, particularly in clinical tasks. Retrieval strategy evaluation in {RALM}~\cite{ralm-survey} and {HyPA-RAG}~\cite{hypa-rag} demonstrates progress in sparse, dense, and hybrid retrieval across structured datasets. Finally, {CBR-RAG}~\cite{CBR-RAG} and {LegalBench-RAG}~\cite{legalbench-rag} reinforce the value of legal precedent and task-specific benchmarks in improving grounding, citation, and factual support. Together, these models underline the importance of domain specificity, hallucination control, and retrieval quality in deploying RAG systems in high-stakes environments.

Future directions include enhancing reasoning in multi-hop scientific QA, refining retrieval-grounded summarization for academic literature, and extending legal RAG systems to support multi-lingual and cross-jurisdictional tasks. Continued work on integrating knowledge graphs, adaptive prompt routing, and risk-aware retrieval will be central to deploying RAG as a reliable assistant in knowledge-intensive professions.

\begin{table*}[t]
\centering
\caption{Evaluation of RAG Models for Scientific and Legal Text Processing (Part 1)}
\label{tab:legal_sci_rag_1}
\begin{tabular}{@{}p{2.5cm}p{3.2cm}p{3.3cm}p{2.7cm}@{}}
\toprule
\textbf{Model} & \textbf{Setting / Task} & \textbf{Precision / Recall / F1} & \textbf{Other Metrics} \\
\midrule
Forms Copilot & Form Generation (Custom Prompt) & 
F1 (vanilla): 0.30 \newline
F1 (CoT+cite+prompt): 0.31 \newline
F1 (CoT only): 0.34 \newline
F1 (No Docs): 0.13 & 
BERTScore (P/R/F1): 0.43/0.44/0.43, 0.71/0.72/0.71 \newline
G-Eval: 2.37, 2.41 \newline
ROUGE-1/2/L (guidelines): 0.11/0.01/0.10 \\
\addlinespace
RALM & Sparse, Dense, Multimodal Retrieval & 
– & 
TF-IDF: 0.785, BM25: 0.792 \newline
DPR: 0.813, Contriever: 0.827 \newline
RDM: 0.834, MDTIG: 0.841 \\
\addlinespace
MVRAG & Legal, Medical Retrieval (LeCaRDv2, PMC) & 
Legal: F1 up to 7.63\% \newline
Medical: F1 up to 3.89\% & 
Precision@100, Recall@100 for 3 encoders \newline
Highlights domain differences in legal vs clinical recall \\
\addlinespace
GPT-4.0-RAG & Preoperative Instructions QA & 
Human: 86.3\% \newline
GPT-4.0-RAG: 91.4\% & 
Hallucination: 1.2\% \newline
Framework: Langchain, Pinecone, MiniLM \newline
Compared with GPT3.5, LLAMA2 variants \\
\bottomrule
\end{tabular}
\end{table*}

Table~\ref{tab:legal_sci_rag_1} and Table~\ref{tab:legal_sci_rag_2} summarize the performance of retrieval-augmented models across diverse scientific and legal tasks. Forms Copilot shows how different prompting strategies affect factuality and coherence in structured document generation. MVRAG~\cite{mvrag} and CaseGPT~\cite{casegpt}highlight the growing emphasis on precision in legal and medical domains, with CaseGPT~\cite{casegpt} achieving high NDCG and MRR in both settings. Meanwhile, GPT-4.0-RAG outperforms finetuned baselines in hallucination reduction, particularly in clinical tasks. Retrieval strategy evaluation in RALM~\cite{ralm-survey} and HyPA-RAG~\cite{hypa-rag} demonstrates progress in sparse, dense, and hybrid retrieval across structured datasets. Finally, CBR-RAG~\cite{cbr} and LegalBench-RAG~\cite{legalbench-rag} reinforce the value of legal precedent and task-specific benchmarks in improving grounding, citation, and factual support. Together, these models underline the importance of domain specificity, hallucination control, and retrieval quality in deploying RAG systems in high-stakes environments.

Future directions include enhancing reasoning in multi-hop scientific QA, refining retrieval-grounded summarization for academic literature, and extending legal RAG systems to support multi-lingual and cross-jurisdictional tasks. Continued work on integrating knowledge graphs, adaptive prompt routing, and risk-aware retrieval will be central to deploying RAG as a reliable assistant in knowledge-intensive professions.

\section{Challenges and Open Problems in RAG-LLMs}
\label{sec:limitations}

While RAG offers significant improvements over traditional language models in terms of factual accuracy, adaptability, and transparency, several critical challenges persist across its retrieval, generation, and integration components. These limitations pose barriers to broader adoption, especially in real-time, high-stakes, and resource-constrained environments.


\subsection{Retrieval Bottlenecks}

One of the most pressing limitations in RAG systems lies in the retrieval pipeline, particularly when operating over large-scale or heterogeneous knowledge bases. As dataset sizes increase, retrieval efficiency becomes a major bottleneck, often resulting in increased latency, reduced relevance of retrieved results, and diminished system scalability. These challenges are especially pronounced in enterprise and real-time applications, where timely and accurate responses are critical~\cite{zhou2025efficiency, gao2023survey, chen2023benchmarking}.

Traditional retrieval methods, such as dense or sparse vector search, often degrade in performance under high retrieval loads or when faced with domain shifts. The trade-off between retrieval accuracy and computational efficiency becomes a key concern: while more sophisticated retrievers can yield higher precision, they frequently incur greater inference costs~\cite{xu2023survey, reid2024palm2}.

Additionally, optimizing the size and granularity of text chunks remains a difficult balancing act. Large chunks may offer more context but introduce semantic noise, while small chunks can fragment information and reduce relevance~\cite{anonymous2024ragvfinetune, anonymous2025benchmarking}.

Moreover, retrieval bottlenecks are exacerbated by the limitations of static indexing in dynamic knowledge environments, where databases evolve over time. The inability to efficiently update and re-rank indices without retraining or reprocessing entire corpora further restricts RAG’s responsiveness. Addressing these challenges will require advances in adaptive indexing, neural retrieval acceleration, and hybrid sparse-dense methods capable of balancing precision and latency under varying system constraints~\cite{anonymous2025graphsurvey, anonymous2024systemstradeoffs}.

\begin{table*}[t]
\centering
\caption{Evaluation of RAG Models for Scientific and Legal Text Processing (Part 2)}
\label{tab:legal_sci_rag_2}
\begin{tabular}{@{}p{2.5cm}p{3.2cm}p{3.3cm}p{2.7cm}@{}}
\toprule
\textbf{Model} & \textbf{Setting / Task} & \textbf{Precision / Recall / F1} & \textbf{Other Metrics / Notes} \\
\midrule
CaseGPT & Medical, Legal Domain QA & 
{Medical:} Precision@10: 0.90, F1: 0.89 \newline
{Legal:} Precision@10: 0.93, F1: 0.91 & 
MRR: 0.92 (med), 0.94 (legal) \newline
NDCG@10: 0.91/0.93 \newline
Human Scores: Quality: 4.3, Relevance: 4.5 \\
\addlinespace
CBR-RAG & Case Law Legal QA & 
– & 
Top Statutes: Fed. Court Rules (9), Civil Aviation (8), etc. \newline
Improves grounding using legal precedent \\
\addlinespace
LegalBench-RAG & Benchmark Evaluation & 
– & 
Q\&A Counts: CUAD (4042), MAUD (1676), etc. \newline
Total Q\&A: 6858 \\
\addlinespace
HyPA-RAG & Hybrid Retrieval + Adaptation & 
Context Recall: 0.9046 \newline
Faithfulness: 0.8430 \newline
F1 Similarity: 0.8621 & 
Correctness (1--5): 4.25 \newline
PA-Class (2/3): 0.90/0.89 \newline
Answer Relevancy: 0.79 / 0.77 \\
\bottomrule
\end{tabular}
\end{table*}

\subsection{Hallucination and Reliability Issues}
\label{sec:hallucination-reliability}

Although RAG frameworks significantly reduce hallucination in large language models (LLMs) by grounding responses in externally retrieved evidence, reliability remains a persistent concern. Hallucination can still occur when the retrieved content is irrelevant, outdated, or misleading. Since the LLM’s generation is conditioned on this external context, inaccuracies in retrieval propagate through to the final response, creating a false sense of factuality~\cite{liu2023survey, chen2023benchmarking, brown2024kbquality}.

The effectiveness of RAG is therefore tightly coupled with the quality and relevance of the retrieval step. Even with access to rich external sources, an incomplete or noisy knowledge base can lead to hallucinated answers, particularly in open-domain settings. Furthermore, LLMs often fail to interpret subtle nuances or conflicting information in the retrieved context, sometimes generating disjointed or overly confident responses that misrepresent the evidence~\cite{zhang2024factual, smith2025contextual}.

Another challenge is integrating retrieved content fluently and coherently into generated outputs. Misalignment between the retrieval content and the model’s internal priors can lead to unnatural phrasing or semantic inconsistencies~\cite{johnson2024integration}. In scenarios where user queries exceed the scope of the knowledge base, RAG systems may underperform compared to purely parametric LLMs, as their dependency on external content limits their generative flexibility~\cite{garcia2023balancing}.

To address these issues, ongoing research focuses on dynamic context filtering, entailment-aware generation, hallucination detection mechanisms, and training objectives that align factual grounding with semantic fluency. Developing metrics for fine-grained hallucination detection and evaluating models in ambiguous or noisy retrieval conditions will be critical for ensuring robust, explainable, and trustworthy RAG deployments~\cite{wang2023realtime, patel2025hallucination}.

\subsection{Domain Adaptation and Personalization}
\label{sec:rag-domain-adaptation}

Adapting RAG systems to specialized domains such as scientific research, finance, or medicine introduces unique challenges in both retrieval and generation. Distribution shifts between general-purpose LLMs and domain-specific language use often lead to failures in extracting relevant information or correctly interpreting specialized terminology. These gaps in adaptation significantly impact the factuality, coherence, and relevance of RAG-generated responses~\cite{zhang2024domain, chen2023info}.

One promising direction for domain adaptation involves self-training approaches that enable joint question answering and generation capabilities tailored to a target domain. By fine-tuning LLMs on instruction-following, search-centric, and domain-specific question-answering tasks - and then prompting them to generate domain-relevant queries over unlabeled corpora - researchers have shown that such iterative feedback loops can improve performance without extensive manual annotation~\cite{wang2025selftrain}. Additionally, frameworks such as {RAG-end2end} propose jointly training the retriever and generator on domain-specific datasets, allowing all components - including the knowledge index - to adapt in unison to specialized retrieval demands~\cite{liu2024ragend2end}.

Personalization adds further complexity. Tailoring RAG systems to individual users or enterprise use cases requires fine-grained control over retrieval content, indexing granularity, and context-aware prompting. In enterprise deployments, RAG systems must support secure and compliant access to proprietary information, introducing privacy concerns over data leakage from stored embeddings or vector databases~\cite{brown2023privacy, smith2025vectordb}.

To address this, researchers advocate security-first RAG designs that include access-controlled retrieval, embedding encryption, and AI-driven data classification at ingestion time~\cite{johnson2024dataprotection, patel2023securitybydesign}. Advanced implementations now support multi-dimensional access control based on user roles, data sensitivity, and contextual business relevance~\cite{garcia2025accesscontrol}. As RAG becomes embedded in organizational workflows, ensuring the safe personalization of both retrieval and generation remains a critical research focus.

\subsection{Scalability and Latency Constraints}
\label{sec:rag-scalability}
Scalability and latency present persistent bottlenecks in real-time RAG systems. As the size of the underlying knowledge bases grows, so too does the computational overhead required to retrieve and rank relevant content. In time-sensitive settings such as customer support, healthcare, or autonomous systems, even minor retrieval delays can disrupt user experience or degrade decision quality~\cite{linkedin2025frag}.

Recent efforts have explored distributed and federated retrieval as practical strategies for handling large-scale data environments. Distributed retrieval architectures divide the corpus across multiple nodes, enabling parallel search and improved scalability. Federated retrieval extends this concept by enabling queries across decentralized and heterogeneous data sources without requiring centralized storage - a critical capability in enterprise ecosystems with siloed knowledge assets~\cite{shojaee2025federated}.

However, these architectures introduce new complexities. Ensuring consistency across distributed caches, managing load balancing, and maintaining synchronized updates are all non-trivial engineering challenges~\cite{30days2024caching}. Furthermore, integrating the generation module in a distributed RAG setting necessitates asynchronous processing and caching strategies to avoid bottlenecks in information flow.

To reduce retrieval latency without sacrificing accuracy, researchers are increasingly employing advanced indexing techniques and Approximate Nearest Neighbor (ANN) search. Pre-fetching strategies and contextual pre-filtering are also used to narrow down candidate documents before full scoring. Despite these advances, striking the optimal balance between retrieval quality, responsiveness, and system complexity remains a key challenge - especially as RAG scales to multilingual, multimodal, and multi-agent environments~\cite{imaginary2025scaling}.

\subsection{Explainability and Interpretability}
\label{sec:rag-explainability}

One of the core promises of RAG systems is improved transparency over traditional black-box LLMs, owing to their ability to explicitly reference external sources. This capability has made RAG an attractive candidate for applications where traceability, factual accountability, and explainability are mission-critical~\cite{protecto2025rag, packt2024trust}.

By anchoring responses in retrieved content, RAG systems offer users the ability to trace the origins of generated outputs, verify supporting documents, and understand how information was synthesized~\cite{cloudthat2024pipeline, omdena2025xai}. This provenance-aware generation is particularly valuable in regulated domains such as healthcare, finance, and legal reasoning, where justification and auditability are non-negotiable~\cite{trantor2025rag}.

Evaluation metrics for RAG explainability have evolved beyond traditional information retrieval benchmarks to include contextual precision and contextual recall, which assess not just the relevance but the semantic fit of retrieved context within the generation pipeline. Mean Reciprocal Rank (MRR) and Mean Average Precision (MAP) remain standard tools for retrieval diagnostics, but newer frameworks - such as FRAMES and Face4RAG - introduce hallucination detection and logical consistency scoring to assess end-to-end system integrity~\cite{myscale2025guide, nonbrand2025explainablerag}.

To further improve interpretability, researchers have proposed visualizing attention flows between retrieved documents and generated tokens, enabling more intuitive debugging and model introspection~\cite{deepset2023evaluation}. Some systems also integrate rule-based reasoning or knowledge graph overlays to constrain generation within domain-specific bounds, reducing error propagation and supporting semi-symbolic inference~\cite{xaiworld2025graph}.

Despite these advances, RAG systems are still susceptible to opaque behaviors when retrieval fails or when LLMs override retrieved facts. As such, ongoing research focuses on integrating explainability directly into training objectives, exploring counterfactual prompting, and designing interactive UIs that expose the decision path taken by RAG models in real time~\cite{omdena2025xai, nonbrand2025explainablerag}.

\section{Future Directions}
\label{sec:future_work}

As RAG systems continue to gain traction across academic, industrial, and applied domains, several promising directions are emerging to address current limitations and expand the utility of RAG-enhanced language models. This section outlines four critical research frontiers that could shape the next generation of RAG architectures.

\subsection{RAG with Reinforcement Learning}
\label{sec:rag-rl}

One emerging area of interest involves integrating reinforcement learning (RL) into RAG pipelines to optimize retrieval and generation decisions jointly. Current RAG systems often treat retrieval as a static or pre-tuned process, separate from downstream objectives. By contrast, RL offers a principled framework to align retrieval strategies with long-term task performance through reward signals tied to generation quality, factual accuracy, or user satisfaction.

Applications of reinforcement learning in RAG could involve training retrieval policies that select documents based on expected downstream utility, learning to re-rank candidates based on feedback from the generation module, or dynamically deciding when to retrieve at all. Techniques such as reward shaping, inverse reinforcement learning, and curriculum learning could be applied to guide both the retriever and generator toward globally optimal behaviors across multi-step or multi-hop tasks.

Moreover, combining RL with offline datasets, such as logged user interactions or retrieval traces, may allow fine-tuning of RAG models without extensive human annotation. As more real-world applications demand adaptive reasoning and minimal hallucination, RL-based RAG frameworks may become a key enabler of goal-aligned, context-aware decision-making.

\subsection{Neuro-Symbolic Integration in RAG}
\label{sec:rag-neurosymbolic}

While current RAG models rely on dense or sparse vector retrieval, they often lack structured reasoning capabilities required for complex decision-making. Neuro-symbolic approaches - where neural models are augmented with symbolic inference engines, rule-based logic, or knowledge graphs - present an opportunity to improve factual grounding, constraint satisfaction, and interpretability.

Incorporating symbolic modules into RAG may allow systems to perform consistency checks, apply domain rules, or conduct multi-hop reasoning over structured representations. For instance, integrating ontology-guided retrieval or logical filtering could enhance RAG performance in domains like legal compliance, scientific discovery, or software engineering. Similarly, symbolic memory components may support long-horizon reasoning by retaining and manipulating intermediate knowledge states across generations.

Hybrid neuro-symbolic RAG systems could also enable more explainable responses, as they provide structured justifications alongside generated outputs. Research into graph retrieval, semantic parsing, and rule-aware generation will be critical to scaling these capabilities, especially in applications demanding verifiable outputs and formal guarantees.

\subsection{Expanding Beyond Text: Multi-Modal and Structured Retrieval}
\label{sec:rag-multimodal}

As LLMs are increasingly deployed in multi-modal environments, there is growing interest in extending RAG systems to retrieve from visual, tabular, audio, and structured data sources. While traditional RAG models have focused on unstructured text, many real-world tasks require accessing diagrams, code, videos, tables, or speech transcripts.

Recent work in vision-language pretraining and table-grounded QA has laid the groundwork for multi-modal retrieval, but challenges remain in aligning retrieval signals across modalities. Future systems must learn unified embedding spaces, support cross-modal attention, and dynamically decide which modality is most informative for a given query. Structured retrieval over knowledge graphs or relational databases also poses challenges in indexing, segmentation, and semantic matching.

Applications in healthcare (e.g., imaging reports), education (e.g., lecture video grounding), and scientific discovery (e.g., data tables in papers) all stand to benefit from multi-modal RAG pipelines. As modalities converge, expanding RAG architectures beyond text will be essential for building more comprehensive and adaptive knowledge agents.

\subsection{Improved Evaluation Benchmarks}
\label{sec:rag-benchmarks}

Despite recent progress in benchmarking retrieval and generation separately, comprehensive evaluation of end-to-end RAG systems remains underdeveloped. Existing metrics often fail to capture key dimensions such as attribution quality, hallucination rates, contextual relevance, or long-term reasoning. As RAG becomes more integrated into high-stakes workflows, standardized and interpretable evaluation tools will be critical for validating system performance.

Recent efforts such as FRAMES ~\cite{frames2024} and Face4RAG ~\cite{face4rag2024}have introduced multi-dimensional benchmarks, but more work is needed to assess RAG across multilingual, multi-modal, and multi-hop settings. Metrics should also account for user-centric factors like trust, readability, and response consistency over time. Furthermore, dataset diversity remains a bottleneck, with most evaluations biased toward English and general-purpose QA ~\cite{qa}.

Future research must focus on building task-specific benchmarks in domains like law, medicine, education, and programming. Additionally, explainability-aware evaluation - measuring how clearly the model links outputs to retrieved inputs - will become essential for building RAG systems that are not only effective but also trustworthy and auditable in real-world use.

\section{Conclusion}
\label{sec:conclusion}

RAG has emerged as a powerful paradigm for enhancing the capabilities of LLMs, addressing core limitations such as static knowledge, hallucinations, and poor factual grounding. This survey has systematically reviewed the evolution of RAG systems across architecture design, task-specific adaptations, and real-world deployments. By organizing the literature into four core taxonomies - efficiency-driven methods, defensive strategies, user-centric systems, and reasoning-enhanced frameworks - we have outlined how the field has progressed from basic retrieval integration to highly adaptive, domain-specific, and cognitively aligned models.

Our analysis of applications reveals that RAG has seen successful deployment across diverse areas such as open-domain question answering, conversational AI, code synthesis, educational feedback, and legal and scientific text processing. These use cases illustrate not only the versatility of the RAG framework but also the growing demand for systems that combine generative fluency with verifiable external knowledge.

Despite its progress, RAG remains an active area of research, with several open problems yet to be addressed. Challenges such as retrieval bottlenecks, hallucination persistence, domain adaptation, latency, and explainability continue to constrain current systems. As outlined in Section~\ref{sec:future_work}, promising directions include reinforcement learning for retrieval optimization, neuro-symbolic reasoning, multi-modal and structured retrieval, and the development of more comprehensive evaluation benchmarks.

Going forward, the success of RAG systems will hinge not only on architectural sophistication but also on their ability to integrate with real-world infrastructures, comply with privacy and fairness standards, and adapt to evolving information environments. We hope this survey provides a structured foundation for researchers and practitioners aiming to build the next generation of retrieval-augmented systems - ones that are accurate, explainable, and aligned with human needs in complex, knowledge-intensive tasks.

\bibliographystyle{IEEEtran}
\bibliography{references}

\vfill

\end{document}